\documentclass[journal]{IEEEtran}

\usepackage{multirow, booktabs, color, soul, threeparttable}
\usepackage{colortbl}
\definecolor{hl}{RGB}{0.75,0.75,0.75}
\sethlcolor{hl}
\definecolor{bestgray}{gray}{0.70}
\newcommand{\best}[1]{\cellcolor{bestgray}\textbf{#1}}
\usepackage{graphicx}
\usepackage{subfigure}
\usepackage{amsmath}
\usepackage{algorithm}
\usepackage{algorithmicx}
\usepackage{algpseudocode}

\algrenewcommand{\algorithmiccomment}[1]{\hfill // #1}
\usepackage{stfloats}
\usepackage{placeins}
\usepackage{verbatim}
\usepackage{marvosym}
\usepackage{cite}
\usepackage[normalem]{ulem}
\usepackage{amssymb}
\usepackage{xcolor}
\usepackage{array}
\usepackage{longtable}
\usepackage{pdflscape}
\usepackage{float}
\usepackage{enumitem}

\begin{document}
\title{Decision Variable Analysis-Guided Differentiated Fuzzy Search for Large-Scale Multi-Objective Optimization}

\author{Boxi~Xiao, Hui~Bai, Jinhua~Zheng,~\IEEEmembership{Member,~IEEE}, Yu~Li, and Juan~Zou,~\IEEEmembership{Member,~IEEE}%
\thanks{This work has been submitted to the IEEE for possible publication. Copyright may be transferred without notice, after which this version may no longer be accessible.}%
\thanks{The authors are with the Key Laboratory of Hunan Province for Internet of Things and Information Security, Xiangtan University, Xiangtan 411105, China, and also with the Key Laboratory of Intelligent Computing and Information Processing, Ministry of Education, Xiangtan University, Xiangtan 411105, China.}%
\thanks{Corresponding authors: Jinhua Zheng (e-mail: jhzheng@xtu.edu.cn) and Hui Bai (e-mail: huibaimonky@163.com).}%
}

\markboth{IEEE TRANSACTIONS ON EMERGING TOPICS IN COMPUTATIONAL INTELLIGENCE, VOL. XX, NO. XX, MONTH YEAR}
{Xiao et al.: Decision Variable Analysis-Guided Differentiated Fuzzy Search for Large-Scale Multi-Objective Optimization}

\maketitle

\begin{abstract}

Large-scale multi-objective optimization problems (LSMOPs) are challenging due to their high-dimensional decision spaces. Fuzzy search is an effective technique for improving search efficiency, while decision variable analysis can reveal the distinct roles of variables in promoting convergence and maintaining diversity. However, existing fuzzy search methods generally employ a uniform search granularity for all variables, overlooking the heterogeneous search requirements implied by variable roles. To address this limitation, this paper proposes a Decision variable analysis-guided Differentiated Fuzzy Search method, termed DDFS.
The proposed method establishes an explicit mapping between decision-variable roles and fuzzy search granularities. Decision variable analysis is employed to identify variable roles and search sensitivities, enabling different variable groups to adopt differentiated fuzzy search behaviors during offspring generation. Furthermore, a Dual-Indicator Stage Transition Mechanism is developed to dynamically adjust fuzzy-updating intensity throughout the evolutionary process, balancing early-stage search-space compression and late-stage convergence refinement.
Extensive experiments on the LSMOP and UF benchmark suites with up to 1000 decision variables show that DDFS generally achieves competitive performance against several representative large-scale multi-objective evolutionary algorithms. The results suggest that explicitly incorporating decision-variable roles into fuzzy search can help improve optimization performance in high-dimensional decision spaces.

\end{abstract}


\IEEEpeerreviewmaketitle

\section{Introduction}
\label{Introduce}

Multi-objective optimization problems (MOPs) widely arise in many real-world applications, such as engineering design~\cite{pereira2022review}, system control~\cite{chen2023balancing}, and resource scheduling~\cite{guo2026scalable}. The goal of these problems is to obtain a set of Pareto-optimal solutions that achieve both convergence toward the Pareto front and diversity along the front. Multi-objective evolutionary algorithms (MOEAs) have been widely used for solving MOPs because population-based search can approximate multiple trade-off solutions in a single run~\cite{zitzler1999spea,knowles1999paes,zitzler2004ibea,beume2007smsemoa,zhou2011survey}. 
As the number of decision variables increases to hundreds or thousands, MOPs become large-scale multi-objective optimization problems (LSMOPs), which are much more difficult to solve due to the sharply enlarged decision space and more complex variable interactions~\cite{LSMOP,LMF,tian2021evolutionary,liu2023survey,liu2024large}. An LSMOP can be formulated as
\begin{equation}
\begin{aligned}
\min\ & \mathbf{F}(\mathbf{x}) = \left(f_1(\mathbf{x}), f_2(\mathbf{x}), \ldots, f_M(\mathbf{x})\right)^{T},\\
\text{s.t.}\ & \mathbf{x}\in\Omega
\end{aligned}
\end{equation}
where $\mathbf{x}=(x_1,\ldots,x_D)$ is a decision vector, $D$ is the number of decision variables, $\mathbf{F}(\mathbf{x})$ is the objective vector, $M$ is the number of objectives, and $\Omega$ denotes the decision space.

The difficulty of LSMOPs mainly stems from the rapid expansion of the decision space as the number of decision variables increases. In high-dimensional decision spaces, the number of potential search directions and variable combinations grows dramatically, whereas the available function-evaluation budget is usually limited~\cite{tian2021evolutionary}. Consequently, only a small fraction of the search space can be explored sufficiently, making promising regions difficult to identify and exploit while increasing the risk of ineffective exploration. Such inefficient search not only delays convergence toward the Pareto-optimal front but also reduces the opportunity to adequately explore different regions of the search space, thereby degrading the diversity of the obtained solutions. Therefore, the fundamental challenge of LSMOPs is to improve search efficiency under limited computational resources while simultaneously maintaining both convergence and diversity.

To address these challenges, extensive studies have explored different strategies to improve search efficiency while maintaining convergence and diversity in high-dimensional decision spaces. Although these methods differ in implementation, they generally achieve this goal by reducing the effective search space, allocating computational resources more efficiently, or generating more promising search directions.
Representative approaches include cooperative coevolution and resource-allocation methods, which decompose the original problem into smaller subproblems or assign more computational resources to promising components, thereby improving the utilization of the limited evaluation budget~\cite{CCGDE3,miguel2016decomposition,CC2}. Problem-transformation and dimension-reduction methods alleviate the curse of dimensionality by reformulating the optimization problem or constructing a lower-dimensional search space~\cite{WOF,ordered,LSMOF,LSMOEA-RD}. Search-strategy-based methods improve offspring generation directly in the original decision space by designing more effective search operators, such as competitive learning~\cite{CSO,LMOCSO,E-CSO}, directed sampling~\cite{LMOEA-S2D}, dual-space search~\cite{LM-DAS}, and adaptive variation strategies~\cite{DGEA}. Despite their different implementations, these approaches mainly regulate where to search, how computational resources should be allocated, or which search directions should be preferred. Comparatively little attention has been paid to how finely different decision variables should be searched during offspring generation.

From the perspective of offspring generation, search granularity is another important factor affecting search efficiency in high-dimensional optimization. Coarse-grained updates can rapidly compress the search space and reduce ineffective exploration, whereas fine-grained updates are essential for accurately exploiting promising regions. Fuzzy search provides an effective mechanism for controlling search granularity by mapping continuous decision variables onto discrete fuzzy levels during the evolutionary process. By restricting the admissible candidate values, fuzzy search effectively narrows the search range and improves search efficiency in high-dimensional decision spaces~\cite{FDV,FDVDS2023}. However, existing fuzzy search methods generally employ a uniform search granularity for all decision variables within the same evolutionary stage.
Such a uniform treatment overlooks the heterogeneous roles of decision variables in LSMOPs. Variables mainly responsible for maintaining diversity often benefit from broader search coverage, whereas variables closely related to convergence require more accurate and conservative refinement. Decision variable analysis~\cite{MOEA/DVA} provides an effective way to identify these heterogeneous variable roles. Nevertheless, existing studies mainly utilize the obtained variable information for variable grouping, search ordering, or computational resource allocation~\cite{LMEA,LERD,WOF-PAG}, rather than explicitly transforming it into variable-wise search granularity control. 
Consequently, the relationship between decision-variable roles and fuzzy search granularity remains largely unexplored, motivating the differentiated fuzzy search framework proposed in this paper.

To address this issue, this paper proposes a Decision variable analysis-guided Differentiated Fuzzy Search (DDFS) method for large-scale multi-objective optimization. By integrating decision variable analysis with fuzzy search, DDFS enables different variable groups to adopt differentiated search granularities during offspring generation according to their optimization roles and search sensitivities. Furthermore, a dual-indicator stage transition mechanism is introduced to dynamically adjust fuzzy-updating intensity throughout the evolutionary process, balancing rapid search-space compression in the early stage with stable convergence refinement in the later stage.
The main contributions of this paper are summarized as follows.
\begin{enumerate}
    \item A decision variable analysis-guided differentiated fuzzy search framework is proposed. The proposed framework establishes an explicit mapping between decision-variable roles and fuzzy search granularities, enabling variable-wise differentiated search during offspring generation.
    \item A sensitivity-aware fuzzy refinement strategy is developed. By further distinguishing convergence-related variables according to their objective-response sensitivities, the proposed strategy adaptively assigns different fuzzy granularities to improve convergence accuracy while maintaining search efficiency.
    \item A dual-indicator stage transition mechanism is designed. The proposed mechanism jointly exploits convergence improvement and decision-space dispersion to dynamically regulate fuzzy-updating intensity, thereby balancing early-stage search-space compression and late-stage convergence refinement.
\end{enumerate}

The remainder of this paper is organized as follows. Section~\ref{Related_work} reviews the related work and presents the motivation of the proposed method. Section~\ref{Proposed_algorithm} describes the proposed DDFS algorithm in detail. Section~\ref{Experimental_results} presents the experimental studies and analyzes the effectiveness of the proposed method.

\section{Related Work}
\label{Related_work}

To improve the search efficiency of MOEAs for LSMOPs, extensive studies have been conducted from different perspectives. Among them, fuzzy search and decision variable analysis are the two research directions most closely related to the proposed DDFS. Therefore, this section first briefly reviews representative MOEA-based approaches for LSMOPs and then discusses fuzzy search and decision variable analysis, with an emphasis on their limitations and the motivation for the proposed method.

\subsection{MOEAs for LSMOPs}
\label{subsec:moeas_lsmops}

Representative MOEAs have achieved remarkable success on conventional MOPs by combining effective offspring generation with environmental selection~\cite{deb2014nsgaIII,cheng2016rvea}. To address the challenges of large-scale multi-objective optimization, numerous MOEA-based approaches have been developed to improve search efficiency while maintaining convergence and diversity.

Existing MOEA-based methods for LSMOPs mainly improve search efficiency from three perspectives. The first category organizes the search process through cooperative coevolution, decision variable analysis, and computational resource allocation. Representative studies decompose the original problem into smaller variable groups, identify variable roles, or allocate more computational effort to promising components, thereby improving the utilization of the limited evaluation budget~\cite{MOEA/DVA,LMEA,LERD,CCGDE3,CC2,DGVI2025}. Although these methods can effectively organize high-dimensional search, their performance usually depends on the quality of variable grouping, role identification, or resource allocation, and inaccurate organization may weaken search efficiency.
The second category reduces the effective search difficulty by reformulating the optimization problem or constructing lower-dimensional search representations, allowing the optimizer to perform search in a more compact decision space~\cite{WOF,LSMOF,APT,IBde2026}. These methods can substantially reduce the search burden when the transformed representation preserves the essential characteristics of the original problem. However, inaccurate transformations or insufficient representations may discard useful search information and consequently degrade optimization performance.
The third category directly improves offspring generation in the original decision space through more effective search strategies, such as competitive learning, directed sampling, adaptive variation, and dual-space search, so that promising search directions can be exploited more efficiently~\cite{LMOCSO,LMOEA-S2D,DGEA,LM-DAS}. Since these methods rely heavily on the quality of the generated search guidance, inaccurate search directions or biased guidance information may reduce their effectiveness during evolution.

Overall, existing studies mainly improve LSMOP performance by organizing variables, reducing the effective search space, allocating computational resources, or guiding search directions. Comparatively little attention has been paid to controlling the search granularity of individual decision variables during offspring generation. Since offspring generation directly determines how candidate solutions are explored in the decision space, search granularity provides another important yet relatively underexplored perspective for improving search efficiency in LSMOPs. Therefore, the following subsections review fuzzy search and decision variable analysis, which are the two research directions most closely related to the proposed method.

\subsection{Fuzzy Search}
\label{subsec:fuzzy_search}

From the perspective of offspring generation, search granularity plays an important role in balancing exploration and exploitation during high-dimensional optimization. Fuzzy search provides an effective mechanism for regulating search granularity while preserving the original decision space. Specifically, it projects the continuous decision-variable values of newly generated offspring onto nearby representative values defined by the current fuzzy granularity, thereby restricting the admissible search positions of individual variables. As a result, the effective search space is compressed, reducing ineffective exploration and improving search efficiency under a limited evaluation budget~\cite{FDV,FDVDS2023}.

FDV is a representative fuzzy search framework for LSMOPs~\cite{FDV}. It divides the evolutionary process into fuzzy evolution and precise evolution. During fuzzy evolution, offspring generated by the embedded MOEA are projected onto nearby representative values according to the current fuzzy granularity before evaluation, enabling the population to search in a compressed decision space. To progressively balance exploration and exploitation, fuzzy evolution adopts a coarse-to-fine granularity schedule, where coarse granularities accelerate the localization of promising regions and finer granularities improve search accuracy. Once the population approaches promising regions, FDV switches to precise evolution for continuous optimization. This framework demonstrates that search granularity can be effectively regulated in the original decision space without modifying the problem representation.
Building upon FDV, some fuzzy-search variants further explore the combination of fuzzy evolution with additional search guidance to generate more informative guiding solutions during offspring generation~\cite{FDVDS2023}. This line of work indicates that fuzzy search can be combined with search-direction guidance. In contrast, the focus of this paper is not to introduce another direction-guided sampling operator, but to investigate how the fuzzy granularity itself can be differentiated across decision variables according to their optimization roles and sensitivities.

Despite these advances, existing fuzzy search methods still determine fuzzy granularity primarily according to the evolutionary stage, causing all decision variables to share the same search granularity within a given stage. Such a uniform strategy effectively compresses the search space but ignores the heterogeneous search requirements of different decision variables. In addition, the transition from fuzzy evolution to precise evolution is usually controlled by predefined schedules rather than the actual optimization state of the population, limiting the adaptability of fuzzy search during evolution. Therefore, existing fuzzy search methods still lack both variable-wise granularity differentiation and adaptive stage regulation, leaving considerable room for further improvement.

\subsection{Decision Variable Analysis}
\label{subsec:dva_lsmops}

Decision variable analysis (DVA) has been widely used in LSMOPs to understand the structure of high-dimensional decision spaces. Early studies mainly employed DVA for variable grouping, problem decomposition, or dimensionality reduction, aiming to reduce the difficulty of large-scale search by organizing decision variables into more manageable components. More recent studies further exploit DVA to identify the functional roles or importance of variables, revealing that different variables may contribute differently to convergence and diversity.

Representative methods such as MOEA/DVA analyze the control properties and interdependencies of decision variables and use the obtained information to decompose the original problem into different variable groups~\cite{MOEA/DVA}. LMEA further identifies convergence-related and diversity-related variables by observing objective-space responses caused by variable perturbations, showing that variable-role information can help organize large-scale search more effectively~\cite{LMEA}. To reduce the computational cost of DVA, LERD reformulates variable analysis as a binary optimization problem and approximates variable grouping results during evolution~\cite{LERD}.

Overall, existing studies demonstrate that DVA is an effective tool for understanding the structure of LSMOPs and exploiting the heterogeneous characteristics of decision variables. Compared with treating all variables uniformly, DVA can reveal which variables play more important roles in convergence and which contribute primarily to maintaining population diversity, thereby providing valuable guidance for improving search efficiency. Accordingly, the obtained variable information has been widely utilized in variable grouping, problem decomposition, search ordering, strategy selection, and computational resource allocation to better organize the optimization process.

However, the role of DVA remains largely confined to organizing the search process rather than regulating the search behavior itself. In most existing studies, decision-variable information serves as prior knowledge for determining how the optimizer should arrange or allocate the search, while its influence on the actual updating behavior of individual variables during offspring generation is still indirect. In particular, although DVA can identify the heterogeneous optimization roles of decision variables, this information has rarely been exploited to guide variable-wise search granularity. Consequently, an explicit mapping between decision-variable roles and fuzzy search granularity has not yet been established.

\subsection{Motivation}
\label{subsec:motivation}

The above review indicates that fuzzy search and decision variable analysis address search efficiency from two complementary perspectives. Fuzzy search directly regulates search granularity during offspring generation, but existing methods assign the same fuzzy granularity to all decision variables within each evolutionary stage. In contrast, decision variable analysis identifies heterogeneous variable roles and sensitivities, yet the obtained information is mainly used to organize the search process rather than guide variable updating. As a result, the relationship between decision-variable characteristics and search granularity remains largely unexplored.

Motivated by this observation, the proposed DDFS establishes an explicit mapping between decision-variable characteristics and fuzzy search granularity. Variables with different roles and sensitivities are assigned differentiated fuzzy granularities during offspring generation, while a dual-indicator stage transition mechanism dynamically adjusts fuzzy-updating intensity according to the search state. In this way, DDFS combines variable-aware granularity assignment with search-state-aware regulation to improve search efficiency while maintaining convergence and diversity.

\section{Proposed Method}
\label{Proposed_algorithm}

The proposed DDFS implements differentiated fuzzy search by mapping decision-variable characteristics to variable-wise fuzzy granularities. DDFS is implemented as a search module and instantiated with LMOCSO in this study. During evolution, the embedded MOEA first generates candidate offspring in the original decision space. DDFS then updates these offspring with fuzzy granularities determined by variable roles, sensitivities, and the current search stage, while retaining the environmental selection of the embedded MOEA.

\begin{figure*}[t]
    \centering
    \IfFileExists{figures/overall_framework.pdf}{%
        \includegraphics[width=\textwidth]{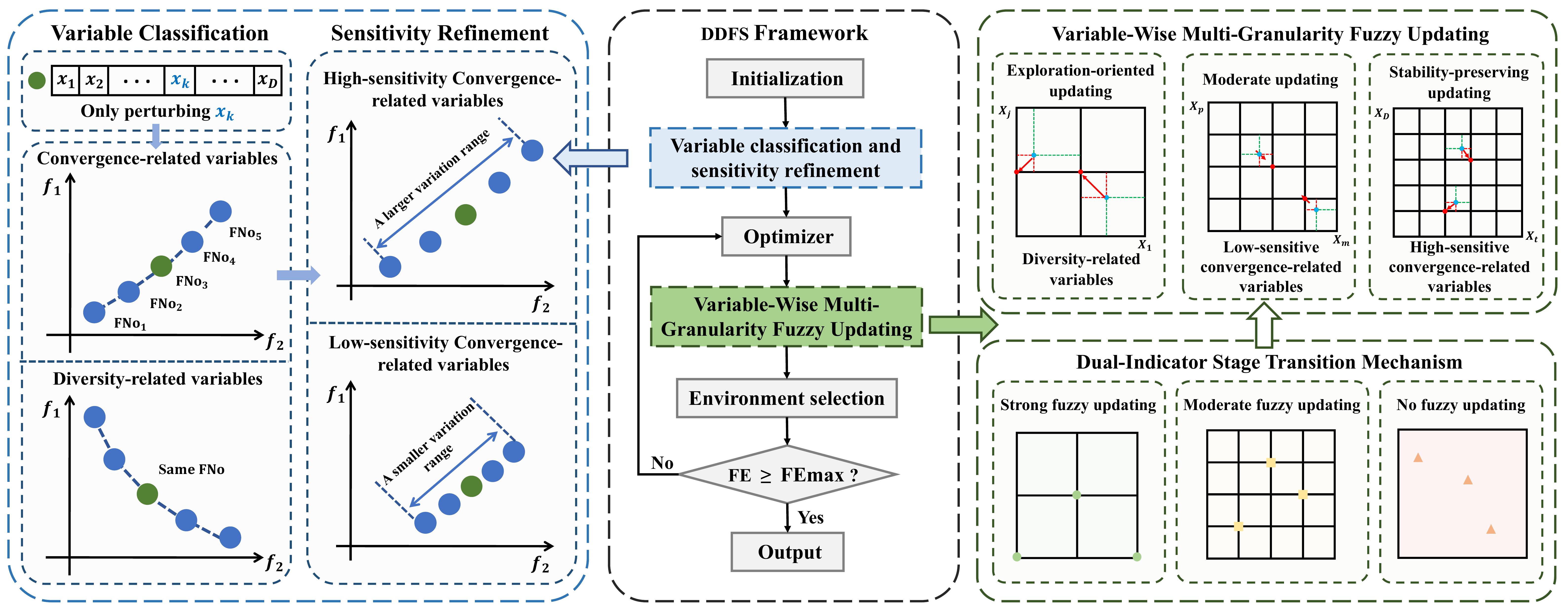}%
    }{%
        \fbox{\parbox{0.95\textwidth}{\centering Missing figure file: figures/overall_framework.pdf}}%
    }
    \caption{Overall framework of DDFS. The variable classification and sensitivity refinement module identifies diversity-related and convergence-related variables, and further divides convergence-related variables into high-sensitivity and low-sensitivity subsets. The base optimizer generates candidate offspring, which are then processed by variable-wise multi-granularity fuzzy updating before environmental selection. The dual-indicator stage transition mechanism determines the fuzzy updating stage during evolution, guiding the progressive transition from strong fuzzy updating to moderate fuzzy updating and finally to no fuzzy updating.}

    \label{fig:overall-framework-ddfl}
\end{figure*}

As illustrated in Fig.~\ref{fig:overall-framework-ddfl} and Algorithm~\ref{alg:overall}, DDFS contains three components. Variable Classification and Sensitivity Refinement identifies diversity-related and convergence-related variables, and further divides the convergence-related variables according to objective-response sensitivity. Variable-Wise Multi-Granularity Fuzzy Updating assigns different grid scales to the obtained variable groups and performs variable-wise fuzzy updating on candidate offspring. Dual-Indicator Stage Transition Mechanism progressively regulates the fuzzy-updating intensity according to the current search state, guiding the transition from strong fuzzy updating to moderate fuzzy updating and finally to disabled fuzzy updating. The computational complexity of DDFS is then analyzed.

\begin{algorithm}[htb]
\caption{Overall Framework of DDFS}
\label{alg:overall}
\small
\begin{algorithmic}[1]
\Require Initial population $P_0$, base MOEA $\mathcal{A}$, maximum function evaluations $FE_{\max}$, number of perturbation samples $N_{\mathrm{PA}}$, historical window length $W$, hysteresis length $K$
\Ensure Final population $P_t$
\State Evaluate $P_0$, set $t=0$, $FE=|P_0|$, and stage label $s_0=1$;
\State Extract the first non-dominated front $F_1(0)$ from $P_0$;
\State $\{I_{\mathrm{div}}, I_{\mathrm{ch}}, I_{\mathrm{cl}}\} \leftarrow \mathrm{VCASR}(P_0,N_{\mathrm{PA}})$; \Comment{Algorithm~\ref{alg:variable_classification}}
\State $FE \leftarrow FE + D N_{\mathrm{PA}}$;
\While{$FE < FE_{\max}$}
    \If{$t > 0$}
        \State $s_t \leftarrow \mathrm{DIST}(P_t, F_1(t-1), s_{t-1}, W, K)$; \Comment{Algorithm~\ref{alg:stage_update}}
    \Else
        \State $s_t \leftarrow s_0$;
    \EndIf
    \State $Q_t \leftarrow \mathrm{OffspringGeneration}_{\mathcal{A}}(P_t)$;
    \State $\widetilde{Q}_t \leftarrow \mathrm{VMFU}(Q_t, \{I_{\mathrm{div}}, I_{\mathrm{ch}}, I_{\mathrm{cl}}\}, s_t)$; \Comment{Algorithm~\ref{alg:fuzzy_update}}
    \State Evaluate the fuzzy-updated offspring $\widetilde{Q}_t$;
    \State $FE \leftarrow FE + |\widetilde{Q}_t|$;
    \State $P_{t+1} \leftarrow \mathrm{EnvironmentalSelection}_{\mathcal{A}}(P_t \cup \widetilde{Q}_t)$;
    \State Extract the first non-dominated front $F_1(t+1)$ from $P_{t+1}$;
    \State $t \leftarrow t + 1$;
\EndWhile
\State \Return $P_t$;
\end{algorithmic}
\end{algorithm}

\subsection{Variable Classification and Sensitivity Refinement}
\label{subsec:variable_classification}

Before the main evolution, DDFS performs variable classification and sensitivity refinement to obtain variable-role priors for fuzzy updating. The variables are first classified into diversity-related and convergence-related sets. The convergence-related variables are then divided into high-sensitivity and low-sensitivity subsets according to their objective-response sensitivities. The resulting sets $I_{\mathrm{div}}$, $I_{\mathrm{ch}}$, and $I_{\mathrm{cl}}$ are used to determine variable-wise fuzzy granularities during offspring updating, where $I_{\mathrm{ch}} \cup I_{\mathrm{cl}} = I_{\mathrm{con}}$.

\begin{algorithm}[htb]
\caption{Variable Classification and Sensitivity Refinement (VCASR)}
\label{alg:variable_classification}
\small
\begin{algorithmic}[1]
\Require Initial population $P_0$, number of perturbation samples $N_{\mathrm{PA}}$
\Ensure Variable prior sets $\{I_{\mathrm{div}}, I_{\mathrm{ch}}, I_{\mathrm{cl}}\}$
\State Randomly select a reference solution $\mathbf{x}^b \in P_0$;
\State Initialize the convergence-related and diversity-related variable sets;
\For{$j = 1$ to $D$}
    \State Generate perturbation samples by perturbing only the $j$-th variable within $[L_j, U_j]$;
    \State Evaluate the perturbation samples and perform non-dominated sorting;
    \If{the maximum front number equals the number of perturbation samples}
        \State Assign the $j$-th variable to the convergence-related set;
    \Else
        \State Assign the $j$-th variable to the diversity-related set;
    \EndIf
\EndFor
\For{each convergence-related variable $j$}
    \State Calculate the objective variation ranges using Eq.~\eqref{eq:variation_range};
    \State Calculate the raw sensitivity score using Eq.~\eqref{eq:sensitivity_score};
\EndFor
\State Normalize the sensitivity scores using Eq.~\eqref{eq:normalized_score};
\State Sort convergence-related variables in descending order of normalized sensitivity;
\State Assign the upper half of the sorted variables to $I_{\mathrm{ch}}$;
\State Assign the lower half of the sorted variables to $I_{\mathrm{cl}}$;
\State \Return $\{I_{\mathrm{div}}, I_{\mathrm{ch}}, I_{\mathrm{cl}}\}$;
\end{algorithmic}
\end{algorithm}

\subsubsection{Variable Classification}

DDFS identifies basic variable roles through single-variable perturbation~\cite{MOEA/DVA,LMEA}. Let $D$ and $M$ denote the numbers of decision variables and objectives, respectively. A reference solution $\mathbf{x}^b \in P_0$ is randomly selected from the initial population. For each variable dimension $j \in \{1, \ldots, D\}$, all other components are fixed, and only the $j$-th variable is uniformly sampled within $[L_j,U_j]$ to generate $N_{\mathrm{PA}}$ perturbation samples:

\begin{equation}
\scalebox{0.89}{$\displaystyle
\mathbf{x}^{(i)} = \left( x_1^b, \dots, x_{j-1}^b, \tilde{x}_j^{(i)}, x_{j+1}^b, \dots, x_D^b \right), \quad i=1,\dots,N_{\mathrm{PA}}.
$}
\label{eq:perturbation_sample}
\end{equation}

The sample set $S_j = \{\mathbf{x}^{(i)}\}_{i=1}^{N_{\mathrm{PA}}}$ is evaluated to obtain the corresponding objective vectors, and non-dominated sorting is performed on $S_j$. Let $\mathrm{MaxFNo}(S_j)$ denote the maximum front number among these samples. If perturbing only the $j$-th dimension produces a clear dominance order, the variable mainly changes the objective values along convergence-related directions and is therefore assigned to the convergence-related set. Specifically, if $\mathrm{MaxFNo}(S_j)=N_{\mathrm{PA}}$, the $j$-th variable is included in $I_{\mathrm{con}}$. Otherwise, the generated samples contain more mutually non-dominated relationships, indicating that the variable is more likely to affect the distribution of solutions in the objective space; in this case, the variable is included in $I_{\mathrm{div}}$. This step gives a coarse division between convergence-related and diversity-related variables, which is refined in the next step.

\subsubsection{Sensitivity Refinement}

The binary distinction between $I_{\mathrm{con}}$ and $I_{\mathrm{div}}$ is insufficient to describe the differences among convergence-related variables. Variables in $I_{\mathrm{con}}$ may still induce different objective responses when perturbed. Therefore, DDFS constructs a sensitivity score from the objective variation ranges of the perturbation samples and refines $I_{\mathrm{con}}$ into $I_{\mathrm{ch}}$ and $I_{\mathrm{cl}}$.

For any $j \in I_{\mathrm{con}}$, the variation range on the $m$-th objective is defined as:
\begin{equation}
    r_{j,m} = \max_{\mathbf{x} \in S_j} f_m(\mathbf{x}) - \min_{\mathbf{x} \in S_j} f_m(\mathbf{x}), \quad m=1, \dots, M.
    \label{eq:variation_range}
\end{equation}

A larger variation range indicates that perturbing only the $j$-th variable can cause a more significant response in the $m$-th objective, implying that this variable is more sensitive with respect to that objective. To characterize the overall sensitivity of a variable over all objectives, the average variation range is used as the raw sensitivity score:
\begin{equation}
    \rho_j = \frac{1}{M} \sum_{m=1}^{M} r_{j,m}.
    \label{eq:sensitivity_score}
\end{equation}

The sensitivity scores are normalized within $I_{\mathrm{con}}$ to obtain comparable values:
\begin{align}
    \rho_{\max} &= \max_{k \in I_{\mathrm{con}}} \rho_k, \\
    \rho_{\min} &= \min_{k \in I_{\mathrm{con}}} \rho_k,
\end{align}
\begin{equation}
    \bar{\rho}_j = \frac{\rho_j - \rho_{\min}}{\rho_{\max} - \rho_{\min}},
    \label{eq:normalized_score}
\end{equation}
After normalization, the variables are sorted in descending order of $\bar{\rho}_j$, and the median is used for binary division. Variables in the higher half are assigned to $I_{\mathrm{ch}}$, while those in the lower half are assigned to $I_{\mathrm{cl}}$. This median-based rule introduces no extra parameter and provides a balanced division across different problems. 

Through the above two-step analysis, DDFS obtains three variable sets $\{I_{\mathrm{div}}, I_{\mathrm{ch}}, I_{\mathrm{cl}}\}$. Here, $I_{\mathrm{div}}$ contains diversity-related variables that mainly affect the distribution and spread of solutions in the objective space; $I_{\mathrm{ch}}$ contains high-sensitivity convergence-related variables that require finer and more conservative fuzzy updating; and $I_{\mathrm{cl}}$ contains low-sensitivity convergence-related variables that can tolerate relatively coarser fuzzy granularity to support steady improvement. These variable categories are obtained once during initialization and used as variable-role priors in subsequent fuzzy updating. This avoids the function-evaluation cost of repeated variable analysis while providing consistent guidance throughout evolution.

\subsection{Variable-Wise Multi-Granularity Fuzzy Updating}

\begin{figure}
    \centering
    \IfFileExists{figures/fuzzy_updating.pdf}{%
        \includegraphics[width=\linewidth]{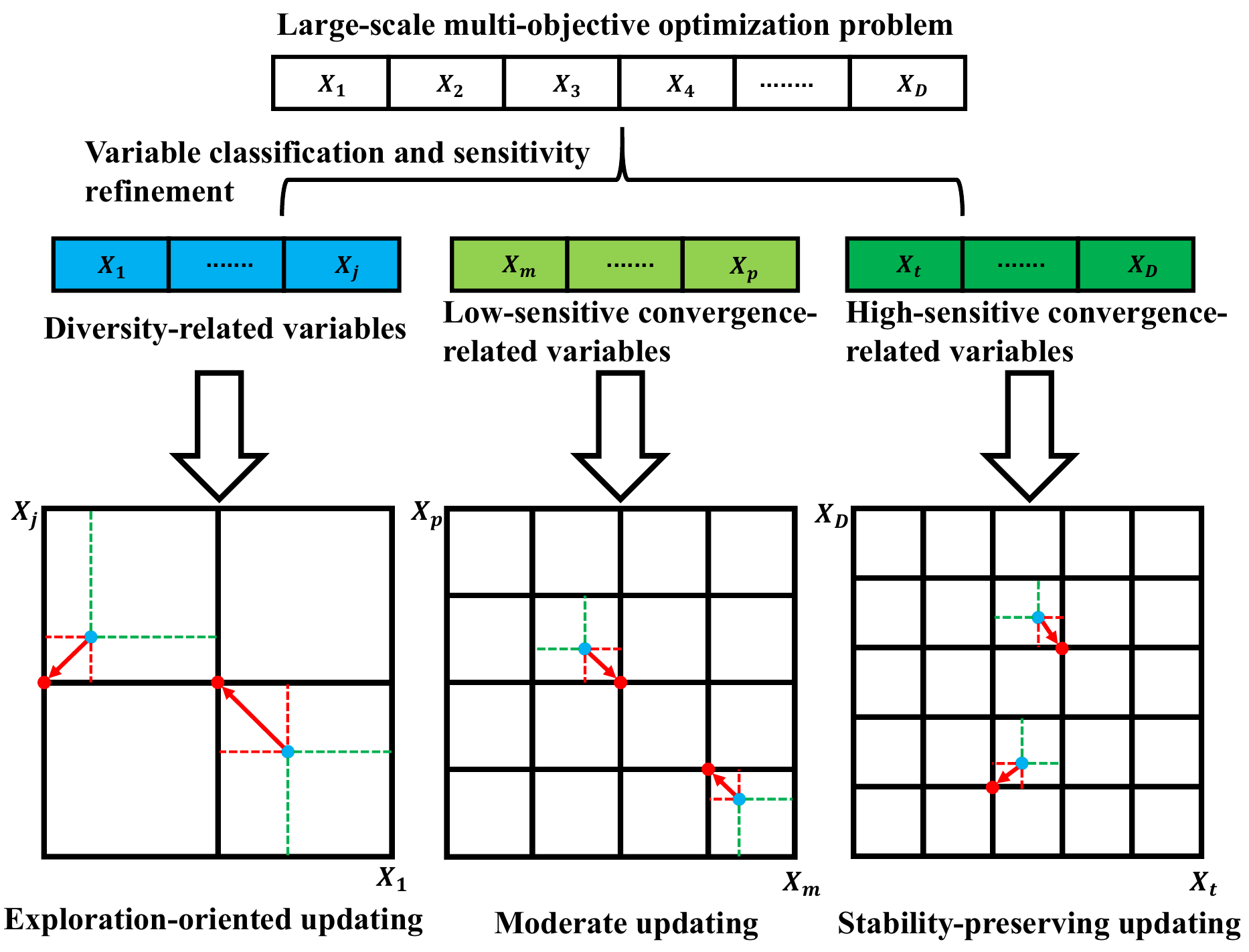}%
    }{%
        \fbox{\parbox{0.9\linewidth}{\centering Missing figure file: figures/fuzzy_updating.pdf}}%
    }
    \caption{Variable-wise multi-granularity fuzzy updating. After variable classification and sensitivity refinement, diversity-related variables, low-sensitivity convergence-related variables, and high-sensitivity convergence-related variables are assigned coarse, intermediate, and fine grid scales, respectively. These settings correspond to exploration-oriented updating, moderate updating, and stability-preserving updating.}
    \label{fig:fuzzy-updating}
\end{figure}

Given $\{I_{\mathrm{div}}, I_{\mathrm{ch}}, I_{\mathrm{cl}}\}$, DDFS applies variable-wise fuzzy updating after the base MOEA generates candidate offspring. This operation does not replace the original offspring-generation procedure. Instead, it maps each offspring variable to a nearby grid node in the original decision space. The grid scale is jointly determined by the variable category and the current fuzzy-updating stage, enabling different variable groups to use different search granularities. As shown in Fig.~\ref{fig:fuzzy-updating}, diversity-related, low-sensitivity convergence-related, and high-sensitivity convergence-related variables correspond to coarse, intermediate, and fine grid scales, respectively.

\subsubsection{Grid-Based Fuzzy Updating}

Consider a candidate solution generated by the base MOEA at generation $t$:
\begin{equation}
    \mathbf{x} = (x_1, x_2, \dots, x_D).
\end{equation}
Let the lower and upper bounds of the $j$-th variable be $L_j$ and $U_j$, respectively. To construct a discrete grid in the normalized space, the variable is first normalized into the interval $[0,1]$:
\begin{equation}
    z_j = \frac{x_j - L_j}{U_j - L_j}, \quad j \in \{1, \dots, D\}.
    \label{eq:fuzzy_normalization}
\end{equation}
Let $C_j$ denote the grid partition number for the $j$-th variable at generation $t$. The corresponding grid step size is defined as:
\begin{equation}
    \delta_j = \frac{1}{C_j}.
    \label{eq:fuzzy_step_size}
\end{equation}

This quantity determines the grid granularity of the $j$-th variable in the normalized space. A larger $C_j$ gives a smaller step size and a finer search scale, whereas a smaller $C_j$ gives a larger step size and a coarser search scale.

In the normalized space, the adjacent lower and upper grid boundaries of $z_j$ are defined as:
\begin{equation}
    z_j^- = \left\lfloor \frac{z_j}{\delta_j} \right\rfloor \delta_j, \quad z_j^+ = \min\{z_j^- + \delta_j, 1\}.
    \label{eq:fuzzy_grid_boundaries}
\end{equation}

The corresponding lower and upper grid points in the original decision space are then calculated as:
\begin{equation}
    y_j^- = L_j + z_j^- (U_j - L_j), \quad y_j^+ = L_j + z_j^+ (U_j - L_j).
    \label{eq:fuzzy_original_grid_points}
\end{equation}

The updated value of the $j$-th variable is selected as the nearer one of these two grid points:
\begin{equation}
    x_j' = 
    \begin{cases} 
        y_j^-, & |x_j - y_j^-| \le |x_j - y_j^+|, \\ 
        y_j^+, & \text{otherwise}. 
    \end{cases}
    \label{eq:fuzzy_nearest_mapping}
\end{equation}

After applying this operation to all decision variables, the fuzzy-updated offspring can be obtained as
\begin{equation}
    \mathbf{x}' = (x_1', x_2', \dots, x_D').
\end{equation}

This process adjusts continuous offspring variables to nearby grid nodes determined by the current fuzzy granularity. After fuzzy updating, the effective candidate values of each variable are restricted to a finite set of grid points rather than arbitrary positions in the continuous interval. For high-dimensional search, this reduces overly fine-grained trials and helps the population search within a more compact set of candidate positions.

\subsubsection{Variable-Category- and Stage-Guided Multi-Granularity Fuzzy Strategy}

In DDFS, variable-wise fuzzy updating is controlled by the grid partition number $C_j$ and the corresponding grid step size $\delta_j=1/C_j$, which defines the fuzzy granularity of the $j$-th variable. For each variable $j$, DDFS determines $C_j$ according to its category, i.e., $j\in I_{\mathrm{div}}$, $j\in I_{\mathrm{cl}}$, or $j\in I_{\mathrm{ch}}$, corresponding to diversity-related, low-sensitivity convergence-related, and high-sensitivity convergence-related variables, respectively, as well as the current stage label $s \in \{1,2,3\}$. In this way, different variables can adopt different grid scales at different search stages.

\begin{algorithm}[htb]
\caption{Variable-Wise Multi-Granularity Fuzzy Updating (VMFU)}
\label{alg:fuzzy_update}
\small
\begin{algorithmic}[1]
\Require Candidate offspring $Q_t$, variable sets $\{I_{\mathrm{div}}, I_{\mathrm{ch}}, I_{\mathrm{cl}}\}$, current stage $s(t)$
\Ensure Fuzzy-updated offspring $\widetilde{Q}_t$
\If{$s(t) = 3$}
    \State \Return $Q_t$; \Comment{Fuzzy updating is turned off in Stage 3}
\EndIf
\State $\widetilde{Q}_t = \emptyset$;
\For{each candidate solution $\mathbf{x} \in Q_t$}
    \State $\mathbf{x}' = \mathbf{x}$;
    \For{$j = 1$ to $D$}
        \State Determine grid partition number $C_j$ based on $s(t)$ and the variable category of $j$;
        \State Calculate the fuzzy step size using Eq.~\eqref{eq:fuzzy_step_size};
        \State Normalize the variable using Eq.~\eqref{eq:fuzzy_normalization};
        \State Calculate the adjacent grid boundaries using Eq.~\eqref{eq:fuzzy_grid_boundaries};
        \State Calculate the grid points in the original decision space using Eq.~\eqref{eq:fuzzy_original_grid_points};
        \State Update the $j$-th variable using Eq.~\eqref{eq:fuzzy_nearest_mapping};
    \EndFor
    \State $\widetilde{Q}_t = \widetilde{Q}_t \cup \{\mathbf{x}'\}$;
\EndFor
\State \Return $\widetilde{Q}_t$;
\end{algorithmic}
\end{algorithm}

In the early search stage (Stage 1), the population is usually far from promising regions and the decision space may contain many low-potential areas. The main purpose of grid-based fuzzy updating is therefore to compress the search range and improve the efficiency of locating promising regions. DDFS adopts relatively strong fuzzy updating: diversity-related variables use coarser grids to maintain broad coverage, high-sensitivity convergence-related variables use finer grids to reduce excessive disturbance, and low-sensitivity convergence-related variables use intermediate grids to support steady improvement.
\begin{equation}
    \delta_j = 
    \begin{cases} 
        \frac{1}{50}, & j \in I_{\mathrm{div}}, \\ 
        \frac{1}{100}, & j \in I_{\mathrm{cl}}, \\ 
        \frac{1}{500}, & j \in I_{\mathrm{ch}}, 
    \end{cases} \quad s = 1.
\end{equation}

In the transition stage (Stage 2), the search shifts from rapid range compression to more stable advancement and refinement. While preserving the relative scale differences among the three variable categories, DDFS further refines the grid scales to reduce unnecessary coarse updates and facilitate stable convergence:
\begin{equation}
    \delta_j = 
    \begin{cases} 
        \frac{1}{500}, & j \in I_{\mathrm{div}}, \\ 
        \frac{1}{1000}, & j \in I_{\mathrm{cl}}, \\ 
        \frac{1}{5000}, & j \in I_{\mathrm{ch}}, 
    \end{cases} \quad s = 2.
\end{equation}

In the late search stage (Stage 3), the population is expected to have approached a relatively stable search region. Continuing fuzzy updating may interfere with fine-grained convergence refinement. Therefore, fuzzy updating is turned off in Stage 3, i.e.,
\begin{equation}
    \mathbf{x}' = \mathbf{x}, \quad s = 3.
\end{equation}

The above grid settings are used as the default configuration in the experiments, and their influence is further examined in the sensitivity analysis.

\subsection{Dual-Indicator Stage Transition Mechanism}
\label{subsec:stage_transition}

\begin{figure*}[t]
    \centering
    \IfFileExists{figures/dual_indicator_stage_transition.pdf}{%
        \includegraphics[
            width=0.95\textwidth,
            height=0.30\textheight,
            keepaspectratio
        ]{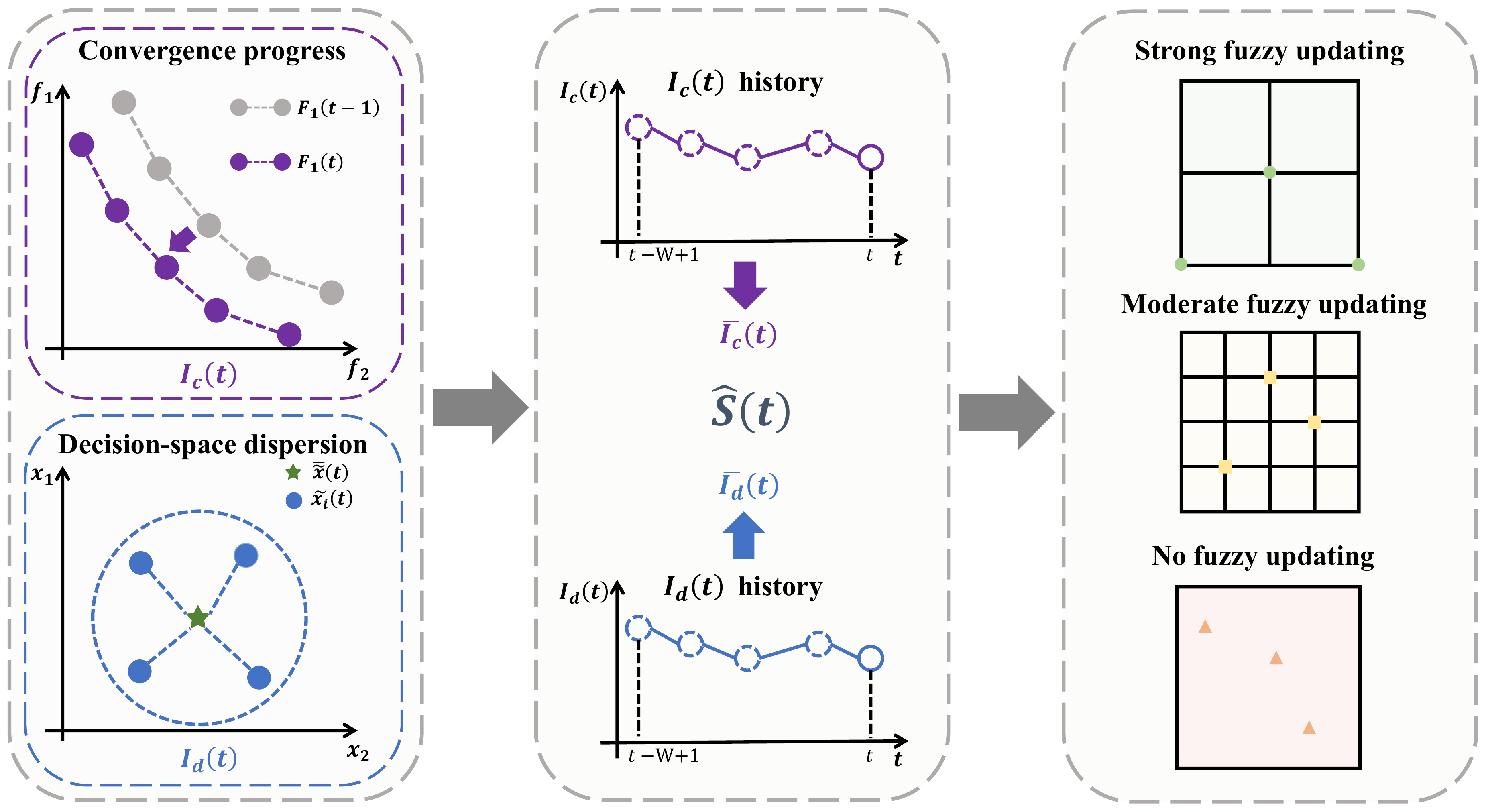}%
    }{%
        \fbox{\parbox{0.95\textwidth}{\centering Missing figure file: figures/dual_indicator_stage_transition.pdf}}%
    }
   \caption{Dual-indicator stage transition mechanism in DDFS. The convergence-improvement and decision-space dispersion indicators are normalized within a sliding window and combined into a composite search-state score, which is used to progressively determine strong fuzzy updating, moderate fuzzy updating, and no fuzzy updating during evolution.}
    \label{fig:dual-indicator-stage-transition}
\end{figure*}

The variable-wise fuzzy updating rules depend on the stage label. Two complementary indicators characterize the current search state: the convergence improvement of the first non-dominated front, which measures whether the population is still making progress in the objective space, and the decision-space dispersion of the population, which measures whether the search region has been sufficiently contracted. By normalizing and combining these indicators, DDFS determines whether fuzzy updating should remain strong, be tightened, or be turned off, as illustrated in Fig.~\ref{fig:dual-indicator-stage-transition}.

\subsubsection{Convergence Improvement}

To characterize convergence improvement, the first non-dominated front at generation $t$ is extracted and denoted as $F_1(t)$. Let the mean value of the $m$-th objective on this front be
\begin{equation}
    \bar{f}_m(t) = \frac{1}{|F_1(t)|} \sum_{\mathbf{x} \in F_1(t)} f_m(\mathbf{x}), \quad m=1, \dots, M,
    \label{eq:front_mean}
\end{equation}
where $M$ is the number of objectives and $|F_1(t)|$ is the number of solutions in the first non-dominated front at generation $t$.

For minimization problems, a smaller mean objective value on the current front than on the previous front indicates improvement along that objective. Based on this, the convergence improvement at generation $t$ is defined as
\begin{equation}
    I_c(t) = \frac{1}{M} \sum_{m=1}^{M} 
    \max \left( 0, 
    \frac{\bar{f}_m(t-1) - \bar{f}_m(t)}
    {|\bar{f}_m(t-1)|} 
    \right),
    \label{eq:conv_improvement}
\end{equation}
A larger $I_c(t)$ means that the current first non-dominated front still shows noticeable improvement, whereas a value close to zero indicates limited convergence progress. The first non-dominated front is used instead of the entire population to reduce the influence of dominated individuals and focus the state evaluation on the leading nondominated solutions.

\subsubsection{Decision-Space Dispersion}

Convergence improvement alone is insufficient to characterize the search state. In high-dimensional decision spaces, limited objective-space improvement does not necessarily mean that the decision-space search region has been sufficiently contracted. Since DDFS directly updates decision variables, a decision-space dispersion indicator is introduced to measure the population dispersion in the normalized decision space.

Let the population size at generation $t$ be $N$, and let the $j$-th decision variable of the $i$-th individual be $x_{i,j}(t)$. To eliminate the influence of different variable ranges, each decision variable is first normalized according to its lower and upper bounds:
\begin{equation}
    \tilde{x}_{i,j}(t) =
    \frac{x_{i,j}(t)-L_j}{U_j-L_j},
    \quad i=1,\dots,N,\; j=1,\dots,D,
    \label{eq:normalized_decision_variable}
\end{equation}
where $L_j$ and $U_j$ are the lower and upper bounds of the $j$-th decision variable, respectively.

The population center in the normalized decision space is then calculated as
\begin{equation}
    \bar{\tilde{x}}_j(t) =
    \frac{1}{N}\sum_{i=1}^{N}\tilde{x}_{i,j}(t),
    \quad j=1,\dots,D.
    \label{eq:normalized_population_center}
\end{equation}

Based on the normalized decision variables, the decision-space dispersion indicator is defined as
\begin{equation}
    I_d(t) =
    \sqrt{
    \frac{1}{ND}
    \sum_{i=1}^{N}
    \sum_{j=1}^{D}
    \left(
    \tilde{x}_{i,j}(t)-\bar{\tilde{x}}_j(t)
    \right)^2
    }.
    \label{eq:decision_dispersion}
\end{equation}

A larger $I_d(t)$ indicates that the population is still widely dispersed and that the decision-space region has not yet been sufficiently contracted. Maintaining stronger fuzzy updating can then help compress the search region and reduce ineffective exploration. In contrast, a smaller $I_d(t)$ indicates that the population has concentrated into a smaller decision-space region, so the need for strong fuzzy updating decreases. In Eq.~\eqref{eq:decision_dispersion}, $ND$ is the total number of normalized decision-variable entries, where $N$ and $D$ denote the population size and the number of decision variables. Thus, $I_d(t)$ reflects the average normalized dispersion over both individuals and dimensions, alleviating scale dependence in high-dimensional spaces. It is used as a decision-space search-state indicator.

\begin{algorithm}[htb]
\caption{Dual-Indicator Stage Transition (DIST)}
\label{alg:stage_update}
\small
\begin{algorithmic}[1]
\Require Population $P_t$, previous front $F_1(t-1)$, previous stage label $s(t-1)$, historical window length $W$, hysteresis length $K$
\Ensure Current stage label $s(t)$
\State Extract $F_1(t)$ from $P_t$;
\State Calculate the average objective values of $F_1(t)$ and $F_1(t-1)$ using Eq.~\eqref{eq:front_mean};
\State Calculate the convergence-improvement indicator using Eq.~\eqref{eq:conv_improvement};
\State Normalize decision variables using Eq.~\eqref{eq:normalized_decision_variable};
\State Calculate the normalized population center using Eq.~\eqref{eq:normalized_population_center};
\State Calculate the decision-space dispersion indicator using Eq.~\eqref{eq:decision_dispersion};
\State Update the sliding window and obtain the max/min values of $I_c(t)$ and $I_d(t)$ within the last $W$ generations;
\State Calculate the normalized indicators $\overline{I}_c(t)$ and $\overline{I}_d(t)$ using Eqs.~\eqref{eq:normalized_conv_indicator} and~\eqref{eq:normalized_disp_indicator};
\State Calculate the composite search-state score using Eq.~\eqref{eq:state_score};
\State Determine the candidate stage using Eq.~\eqref{eq:candidate_stage};
\State Confirm the current stage using the hysteresis mechanism in Eq.~\eqref{eq:confirmed_stage};
\State \Return $s(t)$;
\end{algorithmic}
\end{algorithm}

\subsubsection{Stage Determination and Hysteresis Confirmation}

Because convergence improvement and decision-space dispersion have different scales, DDFS normalizes them before combination. A sliding window of length $W$ is used, and $W$ is set to $5$ in this paper. Before the sliding window is fully filled, all available historical indicator values are used. Before $K$ candidate-stage records are available, the previous stage label is retained. Let the minimum and maximum values of convergence improvement and decision-space dispersion within the most recent $W$ generations be denoted by $I_c^{\min}(t)$, $I_c^{\max}(t)$, $I_d^{\min}(t)$, and $I_d^{\max}(t)$, respectively. The normalized indicators are defined as
\begin{align}
    \overline{I}_c(t) &=
    \frac{I_c(t) - I_c^{\min}(t)}
    {I_c^{\max}(t) - I_c^{\min}(t)},
    \label{eq:normalized_conv_indicator} \\
    \overline{I}_d(t) &=
    \frac{I_d(t) - I_d^{\min}(t)}
    {I_d^{\max}(t) - I_d^{\min}(t)}.
    \label{eq:normalized_disp_indicator}
\end{align}

Based on these two normalized indicators, a composite search-state score is constructed as
\begin{equation}
    S(t) = \alpha \overline{I}_c(t) + (1 - \alpha) \overline{I}_d(t),
    \label{eq:state_score}
\end{equation}
where $\alpha \in [0,1]$ balances objective-space convergence improvement and decision-space dispersion. In this paper, $\alpha = 0.9$, as determined by parameter analysis. A larger $S(t)$ indicates noticeable convergence improvement and high decision-space dispersion, so stronger fuzzy updating is still needed. Conversely, a smaller $S(t)$ indicates that the population is approaching stability in both objective-space improvement and decision-space contraction, so fuzzy updating should be tightened or turned off.

After obtaining the composite state score $S(t)$, two thresholds, $\tau_1$ and $\tau_2$, are used to determine the candidate stage $\hat{s}(t)$:
\begin{equation}
    \hat{s}(t) = 
    \begin{cases} 
        1, & S(t) \ge \tau_1, \\ 
        2, & \tau_2 \le S(t) < \tau_1, \\ 
        3, & S(t) < \tau_2. 
    \end{cases}
    \label{eq:candidate_stage}
\end{equation}

Here, $\hat{s}(t)=1$ denotes the active stage with strong fuzzy updating, $\hat{s}(t)=2$ denotes the transition stage with globally tightened fuzzy updating, and $\hat{s}(t)=3$ denotes the late exploitation stage with fuzzy updating turned off. In this paper, $\tau_1=0.6$ and $\tau_2=0.4$, as selected by parameter analysis.

Directly updating the stage label according to $\hat{s}(t)$ may cause oscillations near the thresholds, because both indicators can fluctuate over successive generations. To avoid repeatedly strengthening and weakening fuzzy updating, a hysteresis-confirmation mechanism is introduced. Let the hysteresis length be $K$. A candidate stage is first accepted as an intermediate stage label $s_c(t)$ only if it remains unchanged for $K$ consecutive generations; otherwise, the previous stage label is retained, i.e.,
\begin{equation}
    s_c(t) = 
    \begin{cases} 
        \hat{s}(t), & \text{if } \hat{s}(t) = \hat{s}(t - 1) = \dots = \hat{s}(t - K + 1), \\ 
        s(t - 1), & \text{otherwise}. 
    \end{cases}
    \label{eq:hysteresis_candidate}
\end{equation}
To ensure progressive stage transition, the stage label is not allowed to move backward once it has advanced:
\begin{equation}
    s(t)=\max\{s(t-1),s_c(t)\}.
    \label{eq:confirmed_stage}
\end{equation}

In this paper, $K=4$, also determined through parameter analysis. This mechanism prevents a temporary threshold crossing from immediately triggering a stage switch unless the change persists. As a result, unnecessary oscillations near the stage boundaries are reduced, and changes in fuzzy-updating intensity become smoother.

\subsection{Complexity Analysis}

The additional cost of DDFS mainly comes from variable analysis, variable-wise fuzzy updating, and stage transition. Variable classification and sensitivity refinement generate $N_{\mathrm{PA}}$ perturbation samples for each of the $D$ variables, resulting in a one-time initialization cost of $O(DN_{\mathrm{PA}})$ additional function evaluations. During evolution, variable-wise fuzzy updating maps each offspring variable to its nearest grid node, costing $O(ND)$ per generation for a population of size $N$. The dual-indicator stage-transition mechanism computes convergence improvement and decision-space dispersion, with an additional cost of at most $O(ND+NM)$ per generation. The memory overhead mainly comes from the three variable index sets and the historical indicator window, approximately $O(D+W)$. Therefore, DDFS introduces a one-time variable-analysis cost and only lightweight per-generation overhead during evolution.

\section{Experimental Results and Analysis}
\label{Experimental_results}

This section evaluates DDFS through three experimental studies. The main manuscript reports the detailed IGD-based comparison and a compact HV-based statistical summary, while the full HV tables are provided in the supplementary material. Section~\ref{subsec:comparison} compares DDFS with representative large-scale multi-objective optimization algorithms. Section~\ref{subsec:ablation_study} presents ablation studies on the main components of DDFS. Section~\ref{subsec:sensitivity_analysis} analyzes the sensitivity of key parameters.

\subsection{Experimental Settings}

\subsubsection{Benchmark Problems and Performance Indicator}

The experiments use two benchmark suites, LSMOP and UF~\cite{LSMOP,UF2009}. LSMOP1--LSMOP9 are used to evaluate the scalability of the algorithms on typical large-scale multi-objective optimization problems, while UF1--UF10 are used to further examine their behavior on problems with different Pareto-front shapes and decision-variable structures. For each test problem, four decision-variable dimensions are considered.

The inverted generational distance (IGD) is adopted as the main performance indicator~\cite{IGD}, and hypervolume (HV) is used as a complementary indicator~\cite{HV}. The main manuscript reports the detailed IGD results and summarizes the HV-based statistical comparison, while the complete HV tables are provided in the supplementary material. IGD can simultaneously measure the convergence and distribution of the obtained nondominated solution set, and a smaller IGD value indicates better overall performance. Let $P^{*}$ be a set of uniformly sampled reference points from the true Pareto front, and let $\Omega$ be the nondominated solution set obtained by an algorithm. IGD is defined as
\begin{equation}
	\mathrm{IGD}(P^{*},\Omega)=\frac{1}{|P^{*}|}\sum_{v\in P^{*}}d(v,\Omega),
    \label{eq:igd_indicator}
\end{equation}
where $d(v,\Omega)$ denotes the Euclidean distance from the reference point $v$ to its nearest solution in $\Omega$.

HV measures the objective-space volume dominated by the obtained nondominated solution set and bounded by a reference point. For minimization problems, let $\mathbf{z}^{r}=(z_1^{r},\ldots,z_M^{r})$ be a reference point dominated by all points of interest, and let $\lambda(\cdot)$ denote the Lebesgue measure. HV is defined as
\begin{equation}
    \mathrm{HV}(\Omega)=\lambda\left(\bigcup_{\mathbf{u}\in\Omega} [f_1(\mathbf{u}),z_1^{r}]\times\cdots\times[f_M(\mathbf{u}),z_M^{r}]\right),
    \label{eq:hv_indicator}
\end{equation}
where a larger HV value indicates better convergence and distribution with respect to the selected reference point.

\subsubsection{Compared Algorithms and Parameter Settings}

DDFS is compared with seven representative large-scale multi-objective optimization algorithms, including FDV~\cite{FDV}, CCGDE3~\cite{CCGDE3}, FLEA~\cite{FLEA}, LMOCSO~\cite{LMOCSO}, LMOEA-S2D~\cite{LMOEA-S2D}, LERD~\cite{LERD}, and DGVI~\cite{DGVI2025}. These algorithms cover important technical categories, such as fuzzy decision variables, cooperative coevolution, decision variable analysis, learning-based search, and resource allocation. Thus, the comparison provides a comprehensive evaluation of whether transforming variable characteristics into differentiated fuzzy search granularities benefits large-scale multi-objective optimization.

For a fair comparison, LMOCSO is adopted as the base MOEA for both the proposed DDFS and the baseline fuzzy search framework FDV, ensuring that any performance discrepancy can be directly attributed to their fuzzy search mechanisms~\cite{LMOCSO}. The population size is set to $N=200$. The number of perturbation samples used in variable classification and sensitivity refinement is set to $N_{\mathrm{PA}}=10$. In the dual-indicator stage-transition mechanism, the convergence-improvement weight is $\alpha=0.9$, the historical window length is $W=5$, the two transition thresholds are $\tau_1=0.6$ and $\tau_2=0.4$, and the hysteresis length is $K=4$. For differentiated fuzzy updating, the grid partition numbers of diversity-related variables, low-sensitivity convergence-related variables, and high-sensitivity convergence-related variables are set to $50, 100, \text{and } 500$ in Stage 1 and $500, 1000, \text{and } 5000$ in Stage 2, respectively. In Stage 3, fuzzy updating is disabled to avoid unnecessary discretization disturbance during late-stage refinement.

All compared algorithms adopt the parameter settings recommended in their original papers or public implementations, and the implementations follow PlatEMO where applicable~\cite{platemo}. Each algorithm is independently run $20$ times on each test instance. The maximum number of function evaluations increases with the decision-variable dimension: $FE_{\max}$ is set to $1,000,000$ for $D=300$, $3,000,000$ for $D=500$, $6,000,000$ for $D=800$, and $8,000,000$ for $D=1000$. The function evaluations consumed by variable classification and sensitivity refinement are counted into the total evaluation budget.

The mean and standard deviation of IGD over $20$ independent runs are reported. Statistical significance is assessed by the Wilcoxon rank-sum test at the significance level of $0.05$ between DDFS and each compared algorithm~\cite{Wilcoxon2013}. In the tables, the symbols ``+'', ``-'', and ``='' indicate that the corresponding compared algorithm is significantly better than, significantly worse than, and statistically similar to DDFS, respectively. The row ``$+/-/=$'' summarizes the numbers of wins, losses, and ties of each compared algorithm against DDFS. The best mean IGD values in the tables are highlighted with gray background and bold font.

\subsection{Comparison with State-of-the-Art Algorithms}

\label{subsec:comparison}
\begin{table*}[!t]
	\centering
	\caption{IGD results obtained by DDFS and the compared algorithms on LSMOP test problems with 300, 500, 800, and 1000 decision variables.}
	\label{tab:lsmop_igd_all_dims}
	\vspace{-1mm}
	\begingroup
	\tiny
	\setlength{\tabcolsep}{1.0pt}
	\renewcommand{\arraystretch}{0.82}
	\resizebox{\textwidth}{!}{%
		\begin{tabular}{ccc|c|c|c|c|c|c|c|c}
			\toprule
			Problem & M & D & FDV & CCGDE3 & FLEA & LMOCSO & LMOEA-S2D & LERD & DGVI & DDFS \\
			\midrule
			\multirow{4}{*}{LSMOP1} & \multirow{4}{*}{2} & 300 & 2.3251e-3 (1.03e-4) - & 3.4100e+0 (2.93e-1) - & 4.1544e-1 (1.27e-1) - & 2.3453e-3 (1.06e-4) - & 3.0007e-1 (9.67e-3) - & 4.4387e-3 (2.65e-4) - & 4.2931e-2 (1.17e-2) - & \best{2.1515e-3 (4.79e-5)} \\
			&  & 500 & 2.0791e-3 (3.78e-5) - & 3.5313e+0 (1.77e-1) - & 3.5063e-1 (1.02e-1) - & \best{1.8188e-3 (1.03e-5) +} & 2.9781e-1 (6.43e-3) - & 3.9563e-3 (1.46e-4) - & 2.0551e-2 (5.54e-3) - & 1.9219e-3 (1.67e-5) \\
			&  & 800 & 2.0640e-3 (4.63e-5) - & 3.8358e+0 (1.47e-1) - & 3.2570e-1 (3.41e-2) - & 2.0091e-3 (2.40e-5) - & 3.0145e-1 (3.06e-3) - & 4.9647e-3 (1.96e-4) - & 2.0926e-2 (5.71e-3) - & \best{1.8275e-3 (8.54e-6)} \\
			&  & 1000 & 2.0649e-3 (4.72e-5) - & 3.9109e+0 (1.27e-1) - & 3.2679e-1 (2.58e-2) - & 2.2281e-3 (3.75e-5) - & 3.0311e-1 (3.38e-3) - & 5.4821e-3 (2.02e-4) - & 2.1499e-2 (4.01e-3) - & \best{1.8106e-3 (1.36e-5)} \\
			\midrule
			\multirow{4}{*}{LSMOP2} & \multirow{4}{*}{2} & 300 & \best{3.5752e-3 (4.73e-4) +} & 1.0343e-1 (1.93e-3) - & 2.7394e-2 (2.79e-3) + & 4.2259e-2 (1.09e-3) - & 1.7846e-2 (2.79e-4) + & 2.0763e-2 (9.39e-3) + & 5.2907e-2 (2.25e-3) - & 3.1311e-2 (4.80e-3) \\
			&  & 500 & \best{2.6066e-3 (1.19e-4) +} & 6.8514e-2 (1.08e-3) - & 1.5976e-2 (2.31e-3) + & 2.2936e-2 (3.11e-4) - & 1.2088e-2 (1.70e-3) + & 7.2989e-3 (2.24e-3) + & 3.7687e-2 (2.09e-3) - & 2.1970e-2 (5.59e-4) \\
			&  & 800 & \best{2.2702e-3 (4.94e-5) +} & 4.7093e-2 (1.20e-3) - & 1.0855e-2 (9.22e-4) + & 1.5856e-2 (1.68e-4) = & 8.8796e-3 (1.09e-3) + & 7.1508e-3 (2.21e-3) + & 2.9905e-2 (9.41e-4) - & 1.5824e-2 (2.93e-4) \\
			&  & 1000 & \best{2.1843e-3 (4.85e-5) +} & 3.8926e-2 (8.22e-4) - & 8.4988e-3 (9.17e-4) + & 1.2900e-2 (1.46e-4) + & 6.7923e-3 (3.79e-4) + & 6.5316e-3 (9.83e-4) + & 2.5366e-2 (4.43e-4) - & 1.3927e-2 (2.52e-4) \\
			\midrule
			\multirow{4}{*}{LSMOP3} & \multirow{4}{*}{2} & 300 & 7.0724e-1 (2.71e-5) = & 1.8128e+1 (1.55e+0) - & 1.5554e+0 (2.35e-3) - & 7.0712e-1 (5.14e-6) = & 1.5531e+0 (1.35e-3) - & 6.7077e-1 (1.80e-2) = & 3.7901e+0 (1.28e+0) - & \best{6.5053e-1 (1.10e-1)} \\
			&  & 500 & 7.0716e-1 (1.04e-5) - & 1.8765e+1 (1.29e+0) - & 1.5644e+0 (1.25e-3) - & 7.0710e-1 (5.00e-5) - & 1.5625e+0 (2.51e-4) - & 8.1692e-1 (5.86e-1) - & 1.5615e+0 (4.18e-1) - & \best{6.5517e-1 (7.02e-2)} \\
			&  & 800 & 7.0715e-1 (5.55e-6) - & 2.0143e+1 (9.52e-1) - & 1.5701e+0 (1.74e-3) - & 7.0711e-1 (2.51e-7) - & 1.5679e+0 (2.10e-4) - & 8.8115e-1 (8.34e-1) - & 1.7267e+0 (4.25e-1) - & \best{6.5083e-1 (4.84e-2)} \\
			&  & 1000 & 7.0715e-1 (4.84e-6) - & 2.2027e+1 (1.40e+0) - & 1.5725e+0 (1.65e-3) - & 7.0711e-1 (2.36e-7) - & 1.5694e+0 (1.74e-4) - & 1.2673e+0 (1.81e+0) - & 1.5726e+0 (2.70e-1) - & \best{6.5760e-1 (4.35e-2)} \\
			\midrule
			\multirow{4}{*}{LSMOP4} & \multirow{4}{*}{2} & 300 & 3.7052e-3 (1.11e-3) - & 1.5078e-1 (6.59e-3) - & 7.1817e-2 (4.97e-3) - & 1.1879e-2 (2.68e-4) - & 6.1236e-2 (1.12e-3) - & 1.0877e-2 (6.01e-4) - & 2.7579e-2 (2.93e-3) - & \best{2.5168e-3 (1.41e-4)} \\
			&  & 500 & 2.3492e-3 (1.88e-4) - & 1.0505e-1 (2.35e-3) - & 4.6233e-2 (4.53e-3) - & 2.2123e-3 (2.72e-4) = & 3.8895e-2 (9.56e-4) - & 7.7982e-3 (7.23e-4) - & 1.2752e-2 (1.78e-3) - & \best{2.0450e-3 (2.84e-5)} \\
			&  & 800 & 1.9872e-3 (4.11e-5) = & 7.3955e-2 (1.23e-3) - & 2.7871e-2 (2.15e-3) - & 2.2990e-3 (1.68e-4) - & 2.5677e-2 (6.66e-4) - & 6.0922e-3 (2.85e-4) - & 1.6160e-2 (4.39e-3) - & \best{1.9670e-3 (2.08e-5)} \\
			&  & 1000 & 1.9348e-3 (4.27e-5) - & 6.2879e-2 (1.25e-3) - & 2.1995e-2 (1.06e-3) - & 1.9103e-3 (9.68e-6) - & 2.0638e-2 (3.50e-4) - & 5.9333e-3 (7.64e-4) - & 1.6256e-2 (5.37e-3) - & \best{1.9071e-3 (4.97e-5)} \\
			\midrule
			\multirow{4}{*}{LSMOP5} & \multirow{4}{*}{2} & 300 & \best{3.2018e-3 (1.50e-4) +} & 7.3201e+0 (1.32e+0) - & 7.4209e-1 (1.15e-16) - & 3.2583e-3 (1.72e-4) + & 7.4209e-1 (1.53e-7) - & 7.0580e-2 (1.11e-1) - & 8.6389e-3 (8.27e-4) - & 3.8755e-3 (1.68e-4) \\
			&  & 500 & 2.8282e-3 (1.12e-4) + & 7.8783e+0 (9.56e-1) - & 7.4209e-1 (1.15e-16) - & \best{2.2112e-3 (4.28e-5) +} & 6.6610e-1 (3.04e-2) - & 2.5768e-2 (8.43e-2) - & 4.9255e-3 (1.95e-4) - & 3.0390e-3 (6.41e-5) \\
			&  & 800 & 2.9557e-3 (1.04e-4) + & 8.4707e+0 (4.14e-1) - & 7.4209e-1 (1.15e-16) - & \best{2.3134e-3 (4.79e-5) +} & 7.4209e-1 (3.32e-7) - & 4.6171e-3 (2.39e-4) - & 4.5648e-3 (3.80e-4) - & 3.6398e-3 (1.35e-4) \\
			&  & 1000 & 3.0242e-3 (8.56e-5) + & 8.7692e+0 (9.55e-1) - & 7.4209e-1 (1.15e-16) - & \best{2.2225e-3 (3.73e-5) +} & 7.4209e-1 (3.99e-7) - & 5.1669e-3 (3.13e-4) - & 4.3584e-3 (9.73e-5) + & 4.4743e-3 (1.84e-4) \\
			\midrule
			\multirow{4}{*}{LSMOP6} & \multirow{4}{*}{2} & 300 & 4.7326e-1 (1.48e-1) - & 8.6392e-1 (4.31e-3) - & \best{1.8879e-1 (1.73e-3) +} & 7.4223e-1 (2.11e-3) - & 2.1733e-1 (2.07e-2) + & 6.7493e-1 (5.74e-2) - & 7.4438e-1 (3.79e-3) - & 2.8363e-1 (1.30e-1) \\
			&  & 500 & 5.4057e-1 (1.14e-1) - & 8.1064e-1 (2.23e-3) - & 2.1261e-1 (9.76e-2) - & 7.3569e-1 (2.03e-2) - & 2.2065e-1 (3.30e-2) - & 6.5827e-1 (9.88e-2) - & 7.4495e-1 (8.00e-4) - & \best{1.3636e-1 (9.28e-2)} \\
			&  & 800 & 6.3845e-1 (2.22e-1) - & 7.8339e-1 (1.24e-3) - & 1.7069e-1 (7.68e-4) - & 6.8402e-1 (9.69e-2) - & 2.2313e-1 (4.54e-2) - & 2.6519e+0 (6.98e+0) - & 7.4848e-1 (1.35e-3) - & \best{6.9099e-2 (2.83e-2)} \\
			&  & 1000 & 7.4392e-1 (3.35e-4) - & 7.5929e-1 (5.83e-2) - & 1.8524e-1 (5.67e-2) - & 6.8765e-1 (9.42e-2) - & 2.2345e-1 (5.32e-2) - & 6.4093e-1 (8.48e-2) - & 7.0458e-1 (1.25e-1) - & \best{5.4967e-2 (1.15e-2)} \\
			\midrule
			\multirow{4}{*}{LSMOP7} & \multirow{4}{*}{2} & 300 & \best{1.0664e+0 (1.26e-1) =} & 1.3916e+4 (3.63e+3) - & 1.4983e+0 (3.24e-5) = & 1.1434e+0 (5.35e-1) = & 1.4986e+0 (2.69e-4) = & 1.7332e+0 (2.49e-1) - & 1.2069e+0 (7.69e-1) = & 1.4372e+0 (5.33e-1) \\
			&  & 500 & 9.6980e-1 (2.26e-1) = & 1.8781e+4 (6.50e+3) - & 1.5060e+0 (8.17e-6) - & 1.1927e+0 (6.07e-1) = & 1.5064e+0 (1.99e-4) - & 1.3745e+0 (1.81e-1) - & \best{2.8563e-1 (6.35e-2) +} & 8.8217e-1 (5.66e-1) \\
			&  & 800 & 9.1180e-1 (1.86e-1) = & 2.0314e+4 (4.50e+3) - & 1.5101e+0 (2.25e-8) - & 1.0337e+0 (6.19e-1) = & 1.5105e+0 (1.49e-4) - & 1.4827e+0 (3.33e-1) - & \best{2.5651e-1 (6.29e-2) +} & 7.8902e-1 (5.07e-1) \\
			&  & 1000 & 9.6232e-1 (1.51e-1) = & 2.1587e+4 (3.19e+3) - & 1.5115e+0 (1.15e-4) - & 1.0359e+0 (6.46e-1) = & 1.5118e+0 (1.73e-4) - & 1.5201e+0 (1.70e-1) - & \best{2.4766e-1 (7.46e-2) +} & 1.0290e+0 (4.11e-1) \\
			\midrule
			\multirow{4}{*}{LSMOP8} & \multirow{4}{*}{2} & 300 & 8.2119e-3 (2.58e-3) - & 6.5414e+0 (8.90e-1) - & 7.4209e-1 (1.15e-16) - & 2.6352e-2 (3.47e-3) - & 6.9640e-1 (4.25e-2) - & 2.6113e-2 (5.87e-3) - & 3.0635e-2 (2.40e-3) - & \best{3.4367e-3 (4.91e-4)} \\
			&  & 500 & 5.4530e-3 (2.43e-3) - & 6.4766e+0 (8.75e-1) - & 7.4209e-1 (1.15e-16) - & 8.5749e-3 (3.70e-3) - & 7.2718e-1 (3.59e-2) - & 1.2746e-2 (4.33e-3) - & 1.0699e-2 (2.22e-3) - & \best{2.6183e-3 (6.17e-5)} \\
			&  & 800 & 3.9887e-3 (1.13e-3) - & 7.0362e+0 (7.66e-1) - & 7.4209e-1 (1.15e-16) - & 6.2714e-3 (1.70e-3) - & 7.3965e-1 (8.56e-3) - & 9.4011e-3 (1.53e-3) - & 6.6384e-3 (5.33e-4) - & \best{2.9453e-3 (1.14e-4)} \\
			&  & 1000 & 3.8235e-3 (1.03e-3) = & 7.7256e+0 (9.39e-1) - & 7.4209e-1 (1.15e-16) - & 3.5722e-3 (1.73e-4) = & 7.4209e-1 (4.61e-7) - & 9.0181e-3 (1.23e-3) - & 5.9372e-3 (5.20e-4) - & \best{3.4961e-3 (6.40e-4)} \\
			\midrule
			\multirow{4}{*}{LSMOP9} & \multirow{4}{*}{2} & 300 & 3.9984e-1 (3.23e-1) - & 1.5495e+1 (4.66e+0) - & 1.6450e-1 (1.79e-1) - & 1.1050e-1 (1.34e-2) - & 3.8172e-1 (3.02e-1) - & 6.5519e-1 (1.72e-1) - & 3.7975e-2 (1.09e-2) - & \best{3.4778e-2 (1.15e-1)} \\
			&  & 500 & 4.8484e-3 (1.26e-4) - & 1.8946e+1 (4.45e+0) - & 6.4499e-2 (6.63e-5) - & 4.0442e-3 (4.69e-5) - & 2.4379e-1 (2.11e-1) - & 4.4593e-1 (6.33e-4) - & 6.1369e-3 (5.93e-4) - & \best{3.9089e-3 (5.95e-5)} \\
			&  & 800 & 5.2320e-3 (1.41e-4) - & 2.4758e+1 (5.03e+0) - & 3.9247e-2 (2.68e-5) - & 6.7221e-3 (2.70e-3) - & 3.9312e-1 (1.90e-1) - & 4.4719e-1 (5.44e-4) - & 5.2449e-3 (4.33e-4) - & \best{3.6597e-3 (3.24e-5)} \\
			&  & 1000 & 6.0538e-3 (3.46e-4) - & 2.2950e+1 (4.79e+0) - & 3.1962e-2 (3.35e-5) - & 3.8670e-3 (9.97e-5) - & 3.0801e-1 (2.10e-1) - & 4.4921e-1 (6.68e-4) - & 5.3947e-3 (5.94e-4) - & \best{3.5799e-3 (1.99e-5)} \\
			\midrule
			$+/-/=$ & & & 8/21/7 & 0/36/0 & 5/30/1 & 6/22/8 & 5/30/1 & 4/31/1 & 4/31/1 & \\
			\bottomrule
		\end{tabular}%
	}
	\endgroup
\end{table*}

\begin{table*}[!t]
	\centering
	\caption{IGD results obtained by DDFS and the compared algorithms on UF test problems with 300, 500, 800, and 1000 decision variables.}
	\label{tab:uf_igd_all_dims}
	\vspace{-1mm}
	\begingroup
	\tiny
	\setlength{\tabcolsep}{1.0pt}
	\renewcommand{\arraystretch}{0.82}
	\resizebox{\textwidth}{!}{%
		\begin{tabular}{ccc|c|c|c|c|c|c|c|c}
			\toprule
			Problem & M & D & FDV & CCGDE3 & FLEA & LMOCSO & LMOEA-S2D & LERD & DGVI & DDFS \\
			\midrule
			\multirow{4}{*}{UF1} & \multirow{4}{*}{2} & 300 & 1.7997e-1 (2.34e-1) - & 7.6415e-1 (6.13e-2) - & 8.3535e-1 (2.43e-1) - & 9.5536e-2 (8.24e-2) - & 2.4907e-1 (4.83e-3) - & 8.7099e-2 (3.71e-2) - & 9.1421e-2 (1.18e-2) - & \best{4.9348e-3 (1.88e-4)} \\
			&  & 500 & 1.7224e-1 (2.28e-1) - & 8.2637e-1 (6.05e-2) - & 8.3973e-1 (2.10e-1) - & 1.9563e-2 (4.01e-2) - & 2.4024e-1 (5.96e-3) - & 6.1716e-2 (7.09e-2) - & 6.1390e-2 (4.01e-3) - & \best{3.1992e-3 (6.42e-5)} \\
			&  & 800 & 8.8860e-2 (1.84e-1) = & 8.6673e-1 (5.08e-2) - & 8.3776e-1 (1.85e-1) - & 2.4278e-2 (5.29e-2) - & 2.4480e-1 (7.52e-3) - & 7.0734e-2 (6.33e-2) - & 6.0765e-2 (8.24e-3) - & \best{3.3664e-3 (8.18e-5)} \\
			&  & 1000 & 1.5715e-1 (2.39e-1) - & 8.9494e-1 (6.12e-2) - & 8.7312e-1 (1.71e-1) - & 3.8096e-3 (1.14e-4) - & 2.4672e-1 (5.41e-3) - & 8.0783e-2 (6.50e-2) - & 6.5291e-2 (7.51e-3) - & \best{3.1706e-3 (1.52e-3)} \\
			\midrule
			\multirow{4}{*}{UF2} & \multirow{4}{*}{2} & 300 & 1.1379e-2 (1.43e-3) - & 3.3335e-1 (5.33e-2) - & 1.2138e-1 (6.50e-3) - & 1.5647e-2 (1.15e-3) - & 8.7791e-2 (1.91e-3) - & 2.6388e-2 (1.24e-3) - & 4.7811e-2 (2.42e-3) - & \best{8.6363e-3 (1.10e-3)} \\
			&  & 500 & 7.7742e-3 (4.33e-4) - & 3.3466e-1 (4.10e-2) - & 1.2184e-1 (7.57e-3) - & 5.5457e-3 (6.65e-4) - & 8.7622e-2 (9.95e-4) - & 2.1460e-2 (1.02e-3) - & 3.6239e-2 (2.05e-3) - & \best{4.9098e-3 (3.64e-4)} \\
			&  & 800 & 6.8803e-3 (4.89e-4) - & 3.7391e-1 (5.43e-2) - & 1.2033e-1 (4.95e-3) - & \best{5.1261e-3 (5.17e-4) +} & 8.7273e-2 (5.74e-4) - & 2.4787e-2 (1.32e-3) - & 3.9658e-2 (2.36e-3) - & 6.2805e-3 (8.85e-4) \\
			&  & 1000 & 6.4465e-3 (4.49e-4) + & 3.7711e-1 (5.07e-2) - & 1.1680e-1 (4.43e-3) - & \best{4.4014e-3 (2.58e-4) +} & 8.7324e-2 (7.45e-4) - & 2.8294e-2 (1.26e-3) - & 4.6683e-2 (2.41e-3) - & 6.9443e-3 (5.68e-4) \\
			\midrule
			\multirow{4}{*}{UF3} & \multirow{4}{*}{2} & 300 & 6.4677e-3 (3.57e-3) = & 4.5643e-1 (5.40e-2) - & 1.5393e-1 (3.58e-3) - & 6.2670e-3 (4.74e-4) - & 1.4209e-1 (2.86e-3) - & 8.9888e-3 (1.37e-3) - & 3.9515e-2 (5.33e-3) - & \best{5.2191e-3 (2.77e-4)} \\
			&  & 500 & 3.8127e-3 (1.60e-3) - & 4.5056e-1 (3.38e-2) - & 1.4143e-1 (2.57e-3) - & 3.1235e-3 (7.72e-5) - & 1.3133e-1 (2.39e-3) - & 4.0784e-3 (3.79e-4) - & 1.9822e-2 (6.21e-3) - & \best{2.4378e-3 (3.26e-5)} \\
			&  & 800 & 4.1836e-3 (1.74e-3) - & 4.8324e-1 (5.35e-2) - & 1.3331e-1 (2.68e-3) - & 3.0882e-3 (1.32e-4) - & 1.2498e-1 (1.19e-3) - & 4.2945e-3 (1.32e-4) - & 1.7023e-2 (4.55e-3) - & \best{2.4939e-3 (4.94e-5)} \\
			&  & 1000 & 3.4445e-3 (1.41e-3) = & 5.0292e-1 (6.05e-2) - & 1.2964e-1 (2.39e-3) - & \best{2.2981e-3 (3.86e-5) +} & 1.2287e-1 (1.03e-3) - & 5.1197e-3 (3.41e-4) - & 1.5117e-2 (3.72e-3) - & 3.3715e-3 (1.38e-4) \\
			\midrule
			\multirow{4}{*}{UF4} & \multirow{4}{*}{2} & 300 & 8.8651e-2 (6.22e-3) - & 1.4125e-1 (4.22e-3) - & 1.2629e-1 (2.26e-3) - & 1.2212e-1 (2.90e-3) - & 5.8514e-2 (1.73e-4) - & 6.8469e-2 (4.65e-3) - & 5.3574e-2 (4.88e-3) - & \best{4.5342e-2 (8.45e-4)} \\
			&  & 500 & 7.6506e-2 (7.10e-3) - & 1.5113e-1 (3.72e-3) - & 1.2854e-1 (1.78e-3) - & 1.1570e-1 (2.00e-3) - & 5.8680e-2 (1.79e-4) - & 5.8429e-2 (2.55e-3) - & 5.7793e-2 (4.11e-3) - & \best{3.9535e-2 (1.12e-3)} \\
			&  & 800 & 7.4021e-2 (6.63e-3) - & 1.5826e-1 (2.50e-3) - & 1.3036e-1 (1.87e-3) - & 1.1894e-1 (1.95e-3) - & 5.8865e-2 (8.87e-5) - & 6.3214e-2 (9.94e-3) - & 6.6823e-2 (1.17e-2) - & \best{3.8386e-2 (1.71e-4)} \\
			&  & 1000 & 6.9427e-2 (5.78e-3) - & 1.6061e-1 (3.91e-3) - & 1.3170e-1 (2.33e-3) - & 1.1752e-1 (1.42e-3) - & 5.8899e-2 (4.40e-5) - & 6.3179e-2 (7.45e-3) - & 6.6027e-2 (3.65e-3) - & \best{3.8647e-2 (5.67e-4)} \\
			\midrule
			\multirow{4}{*}{UF5} & \multirow{4}{*}{2} & 300 & 7.0155e-1 (1.07e-2) - & 3.4456e+0 (2.74e-1) - & 4.3120e+0 (8.89e-1) - & 7.7891e-1 (2.25e-1) - & 2.6485e+0 (1.08e-1) - & 6.1656e-1 (1.43e-1) - & 9.2267e-1 (9.54e-2) - & \best{4.8170e-1 (2.73e-2)} \\
			&  & 500 & 7.0596e-1 (4.03e-3) - & 3.3812e+0 (1.65e-1) - & 3.8918e+0 (7.41e-1) - & 4.0511e-1 (1.24e-1) - & 2.6459e+0 (5.78e-2) - & 5.1298e-1 (1.41e-1) - & 3.5274e-1 (2.49e-2) = & \best{3.3694e-1 (1.68e-2)} \\
			&  & 800 & 6.7194e-1 (1.30e-1) - & 3.7169e+0 (1.69e-1) - & 4.7234e+0 (7.67e-1) - & 4.3489e-1 (1.42e-1) - & 2.7640e+0 (7.35e-2) - & 5.1468e-1 (2.56e-1) - & 3.3979e-1 (1.63e-2) - & \best{2.4599e-1 (9.24e-3)} \\
			&  & 1000 & 6.7489e-1 (1.25e-1) - & 3.7282e+0 (1.90e-1) - & 4.4931e+0 (6.60e-1) - & 4.0952e-1 (3.54e-2) - & 2.8446e+0 (9.03e-2) - & 6.9252e-1 (4.77e-1) - & 3.8120e-1 (2.06e-2) - & \best{2.3188e-1 (5.73e-3)} \\
			\midrule
			\multirow{4}{*}{UF6} & \multirow{4}{*}{2} & 300 & 5.2852e-1 (4.64e-3) - & 3.2598e+0 (4.15e-1) - & 2.8263e+0 (1.12e+0) - & 3.9440e-1 (9.73e-2) - & 1.0700e+0 (5.19e-2) - & 4.2491e-1 (5.38e-2) - & 1.6851e-1 (9.55e-2) - & \best{4.8034e-2 (6.12e-3)} \\
			&  & 500 & 5.3039e-1 (4.02e-4) - & 3.4509e+0 (3.65e-1) - & 2.8914e+0 (1.11e+0) - & 3.2675e-1 (1.28e-1) - & 1.0598e+0 (4.23e-2) - & 4.1200e-1 (1.88e-1) - & 1.8847e-1 (1.11e-1) - & \best{2.0928e-2 (1.16e-2)} \\
			&  & 800 & 5.3046e-1 (4.04e-5) - & 3.7450e+0 (3.24e-1) - & 3.2213e+0 (1.30e+0) - & 3.8490e-1 (8.56e-2) - & 1.0579e+0 (5.54e-2) - & 4.6577e-1 (1.89e-1) - & 1.6393e-1 (8.03e-2) - & \best{1.4343e-2 (6.46e-4)} \\
			&  & 1000 & 5.3031e-1 (5.61e-4) - & 3.5720e+0 (2.72e-1) - & 3.4201e+0 (9.21e-1) - & 3.2680e-1 (1.46e-1) - & 1.0938e+0 (2.59e-2) - & 4.2588e-1 (7.25e-2) - & 1.6862e-1 (9.15e-2) - & \best{1.5295e-2 (4.84e-4)} \\
			\midrule
			\multirow{4}{*}{UF7} & \multirow{4}{*}{2} & 300 & 2.0589e-1 (3.13e-1) - & 8.6672e-1 (6.57e-2) - & 7.9289e-1 (2.47e-1) - & 1.3303e-1 (1.57e-1) - & 2.9616e-1 (9.20e-2) - & 4.2297e-2 (9.63e-3) - & 4.3942e-2 (1.37e-3) - & \best{6.0586e-3 (2.27e-3)} \\
			&  & 500 & 1.1475e-1 (2.41e-1) - & 8.7211e-1 (6.10e-2) - & 8.4098e-1 (3.45e-1) - & 6.1745e-2 (1.19e-1) - & 2.8397e-1 (7.96e-2) - & 3.1983e-2 (4.75e-3) - & 3.5113e-2 (1.01e-3) - & \best{3.2088e-3 (1.26e-4)} \\
			&  & 800 & 2.4788e-1 (3.36e-1) - & 9.2033e-1 (6.31e-2) - & 7.2100e-1 (1.89e-1) - & 3.5313e-2 (1.71e-2) - & 3.2517e-1 (1.02e-1) - & 3.9072e-2 (5.81e-3) - & 3.6270e-2 (9.51e-4) - & \best{3.4249e-3 (1.70e-4)} \\
			&  & 1000 & 2.5170e-1 (3.34e-1) - & 9.5311e-1 (5.51e-2) - & 7.1767e-1 (3.00e-1) - & 6.7962e-2 (1.25e-1) - & 3.9763e-1 (9.41e-2) - & 5.0811e-2 (1.58e-2) - & 3.8688e-2 (7.73e-4) - & \best{3.8682e-3 (2.09e-4)} \\
			\midrule
			\multirow{4}{*}{UF8} & \multirow{4}{*}{3} & 300 & 3.2173e-1 (2.54e-1) - & 1.3544e+0 (1.90e-1) - & 6.3241e-1 (2.95e-2) - & 4.7605e-1 (3.91e-2) - & 4.1760e-1 (4.34e-2) - & 5.2514e-1 (5.94e-3) - & 2.5110e-1 (1.67e-3) - & \best{1.4318e-1 (4.15e-2)} \\
			&  & 500 & 2.9723e-1 (1.79e-1) - & 1.4411e+0 (2.13e-1) - & 6.2031e-1 (6.38e-2) - & 5.3788e-1 (1.36e-2) - & 3.6658e-1 (2.63e-2) - & 5.3046e-1 (5.01e-3) - & 2.4876e-1 (3.08e-3) - & \best{1.0432e-1 (4.09e-2)} \\
			&  & 800 & 3.4204e-1 (2.45e-1) - & 1.5264e+0 (1.79e-1) - & 5.8319e-1 (9.10e-2) - & 5.4215e-1 (1.55e-3) - & 3.6158e-1 (1.55e-2) - & 5.3183e-1 (5.26e-3) - & 2.5956e-1 (9.08e-3) - & \best{1.5227e-1 (5.78e-2)} \\
			&  & 1000 & 3.4264e-1 (2.45e-1) - & 1.5810e+0 (1.96e-1) - & 5.9327e-1 (9.95e-2) - & 5.4022e-1 (6.60e-3) - & 3.6273e-1 (1.31e-2) - & 5.3391e-1 (3.17e-3) - & 2.6679e-1 (1.12e-2) - & \best{1.9935e-1 (1.04e-1)} \\
			\midrule
			\multirow{4}{*}{UF9} & \multirow{4}{*}{3} & 300 & 3.3841e-1 (7.46e-2) - & 1.6360e+0 (2.35e-1) - & 6.6187e-1 (3.61e-2) - & 2.9592e-1 (2.38e-2) - & 5.4438e-1 (2.15e-2) - & 3.3244e-1 (5.74e-2) - & 4.9107e-1 (5.16e-2) - & \best{7.4390e-2 (3.38e-2)} \\
			&  & 500 & 3.8051e-1 (8.04e-2) - & 1.6040e+0 (1.69e-1) - & 6.6186e-1 (3.60e-2) - & 2.8072e-1 (2.80e-2) - & 4.9866e-1 (3.45e-2) - & 3.8556e-1 (1.34e-2) - & 4.5629e-1 (3.81e-2) - & \best{4.7295e-2 (1.32e-2)} \\
			&  & 800 & 3.2825e-1 (1.39e-1) - & 1.6067e+0 (5.74e-2) - & 6.5460e-1 (3.37e-2) - & 2.9420e-1 (1.26e-2) - & 4.9392e-1 (2.62e-2) - & 3.5903e-1 (1.01e-1) - & 4.8467e-1 (5.09e-2) - & \best{6.8513e-2 (1.14e-2)} \\
			&  & 1000 & 3.6885e-1 (1.07e-1) - & 1.6639e+0 (3.36e-2) - & 6.5000e-1 (3.31e-2) - & 2.9794e-1 (1.95e-2) - & 4.8939e-1 (2.19e-2) - & 4.0587e-1 (8.33e-3) - & 5.3849e-1 (1.76e-2) - & \best{7.3700e-2 (9.45e-3)} \\
			\midrule
			\multirow{4}{*}{UF10} & \multirow{4}{*}{3} & 300 & 7.3253e-1 (2.02e-1) - & 7.9452e+0 (4.52e-1) - & 4.1364e+0 (1.59e-1) - & 7.1246e-1 (5.67e-2) - & 3.1218e+0 (1.37e-1) - & 5.6903e-1 (2.03e-1) = & 5.2506e-1 (8.10e-2) = & \best{4.8470e-1 (6.52e-2)} \\
			&  & 500 & 7.1966e-1 (1.96e-1) - & 8.5848e+0 (9.95e-1) - & 4.1277e+0 (2.16e-1) - & 6.4966e-1 (8.03e-2) - & 3.0365e+0 (2.01e-1) - & 5.5522e-1 (9.87e-2) = & 5.1556e-1 (1.41e-1) = & \best{4.8359e-1 (6.47e-2)} \\
			&  & 800 & 5.8346e-1 (2.91e-1) = & 8.6046e+0 (8.36e-1) - & 4.2947e+0 (2.13e-1) - & 6.5383e-1 (6.39e-2) - & 3.1324e+0 (1.06e-1) - & 6.6442e-1 (1.77e-1) = & 5.6850e-1 (2.73e-2) = & \best{5.4679e-1 (4.51e-2)} \\
			&  & 1000 & 6.4030e-1 (2.54e-2) = & 8.8478e+0 (3.48e-1) - & 4.2252e+0 (1.83e-1) - & 6.8897e-1 (4.45e-2) = & 3.2196e+0 (2.50e-1) - & 8.1739e-1 (2.07e-1) - & 6.6816e-1 (2.40e-2) = & \best{5.9209e-1 (2.98e-1)} \\
			\midrule
			$+/-/=$ & & & 1/34/5 & 0/40/0 & 0/40/0 & 3/36/1 & 0/40/0 & 0/37/3 & 0/35/5 & \\
			\bottomrule
		\end{tabular}%
	}
	\endgroup
\end{table*}

Tables~\ref{tab:lsmop_igd_all_dims} and~\ref{tab:uf_igd_all_dims} report the IGD results on the LSMOP and UF test problems, respectively.
An analysis of the overall results shows that the proposed DDFS method obtains the best mean IGD values on a majority of the test instances. Specifically, DDFS achieves the best mean IGD values on 15, 15, 15, and 14 out of the 19 instances for decision-variable dimensions of 300, 500, 800, and 1000, respectively. This result suggests that DDFS remains competitive as the decision-variable dimension increases. Moreover, DDFS shows favorable results on both the LSMOP and UF benchmark suites, indicating that the proposed differentiated fuzzy search can remain applicable to problems with different landscape characteristics and decision-variable structures.

On the LSMOP suite, DDFS also obtains competitive results, achieving the best mean IGD values on 22 out of 36 instances. The improvements are relatively evident on LSMOP1, LSMOP3, LSMOP4, LSMOP6, LSMOP8, and LSMOP9. These results indicate that the proposed variable-wise fuzzy granularity is useful for many large-scale problems, although some compared algorithms remain competitive on several instances. By assigning coarser grids to diversity-related variables, DDFS can maintain broad search coverage and support search-space compression. Meanwhile, by assigning finer grids to high-sensitivity convergence-related variables, DDFS can reduce excessive disturbance along sensitive convergence directions and promote more stable refinement.

It is also worth noting that DDFS does not obtain the best result on every LSMOP instance. For example, on LSMOP2, LSMOP5, and LSMOP7, some compared algorithms still achieve better mean IGD values under certain dimensional settings. This phenomenon indicates that the performance of DDFS may still be influenced by problem-dependent variable interactions and landscape properties. Nevertheless, the overall results on the LSMOP suite show that transforming variable-role information into differentiated fuzzy search granularities is beneficial for improving the search behavior in many high-dimensional decision spaces.

The UF benchmark suite provides additional evidence for the effectiveness of DDFS. DDFS obtains the best mean IGD values on 37 out of 40 UF instances, showing favorable performance on many problems with different Pareto-front shapes and decision-variable structures. Compared with uniform fuzzy search, DDFS reduces the resolution mismatch among different variable categories by assigning different grid scales according to variable roles and sensitivities. In addition, the dual-indicator stage-transition mechanism weakens or disables fuzzy updating when the search becomes relatively stable, which helps reduce unnecessary late-stage discretization disturbance and preserve the quality of the obtained solution set.

\begin{figure*}[!t]
	\centering
	\IfFileExists{figures/uf500_igd_convergence_curves.pdf}{%
		\includegraphics[width=\textwidth]{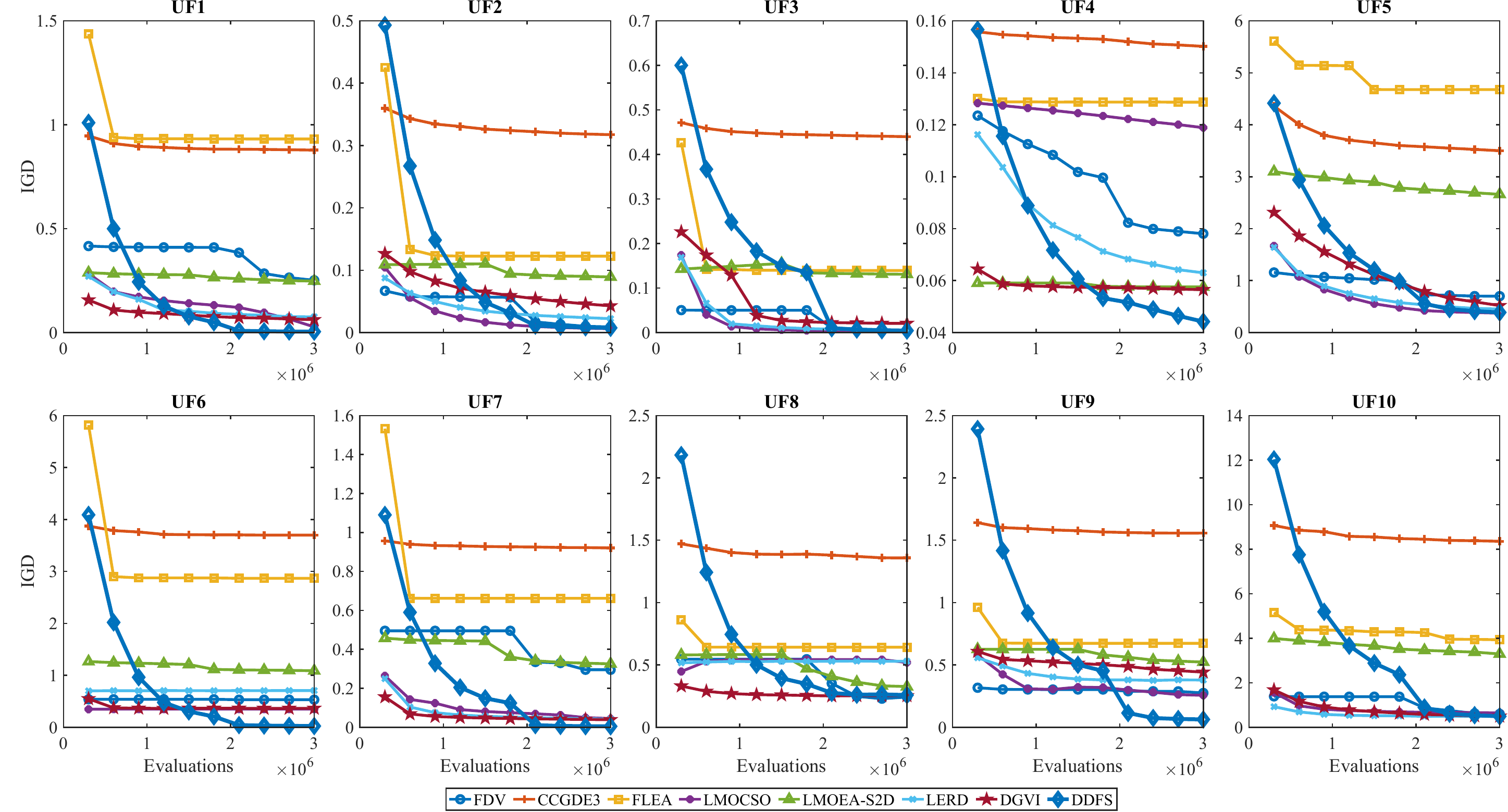}%
	}{%
		\fbox{\parbox{0.95\textwidth}{\centering Missing figure file: \texttt{figures/uf500\_igd\_convergence\_curves.pdf}}}%
	}
	\caption{IGD convergence curves of DDFS and the compared algorithms on UF1--UF10 with $D=500$. The curves report the mean IGD values over 20 independent runs, and a smaller IGD value indicates better performance.}
	\label{fig:uf500_igd_curves}
\end{figure*}

Fig.~\ref{fig:uf500_igd_curves} further illustrates the IGD convergence behavior of DDFS and the compared algorithms on the UF problems with $D=500$. In most UF instances, DDFS shows a rapid decrease in IGD during the early and middle stages and reaches lower final IGD values than the compared algorithms. This observation is consistent with the numerical results in Table~\ref{tab:uf_igd_all_dims}. It indicates that the proposed variable-wise fuzzy updating can help the population approach promising regions efficiently. 
The statistical results further support the competitiveness of the proposed method. Across all 76 instances, DDFS significantly outperforms CCGDE3, FLEA, LMOEA-S2D, LERD, DGVI, LMOCSO, and FDV on 76, 70, 70, 68, 66, 58, and 55 instances, respectively. These results suggest that simply compressing the decision space with a uniform fuzzy granularity or relying only on resource allocation may be insufficient for many high-dimensional multi-objective optimization problems. In contrast, explicitly mapping decision-variable roles and sensitivities to differentiated fuzzy search granularities can help improve the balance between convergence and distribution maintenance.

To further examine whether the above observations are consistent under a volume-based indicator, Table~\ref{tab:hv_summary_main} summarizes the HV-based statistical comparison, and the complete HV values are provided in the supplementary material. On the LSMOP suite, DDFS obtains the best mean HV values on 17 out of 36 instances. Although FDV and LMOCSO remain competitive on several LSMOP instances, DDFS still significantly outperforms CCGDE3, FLEA, LMOEA-S2D, LERD, and DGVI on 34, 29, 30, 28, and 31 instances, respectively. On the UF suite, DDFS obtains the best mean HV values on 34 out of 40 instances and shows clear advantages over most compared algorithms. Overall, DDFS obtains the best mean HV values on 51 out of all 76 instances. These HV results are broadly consistent with the IGD results and provide complementary evidence that the proposed differentiated fuzzy search can improve both convergence and distribution on many high-dimensional test instances, while its advantage remains problem-dependent on several LSMOP cases.

\begin{table*}[!t]
\centering
\caption{HV-based statistical summary of DDFS against the compared algorithms. For HV, a larger value indicates better performance. The symbols ``+'', ``-'', and ``='' denote that the corresponding compared algorithm is significantly better than, significantly worse than, and statistically similar to DDFS, respectively.}
\label{tab:hv_summary_main}
\vspace{-1mm}
\begingroup
\scriptsize
\setlength{\tabcolsep}{4pt}
\renewcommand{\arraystretch}{1.05}
\resizebox{\textwidth}{!}{%
\begin{tabular}{c|c|c|c|c|c|c|c|c}
\toprule
Benchmark & Best HV of DDFS & FDV & CCGDE3 & FLEA & LMOCSO & LMOEA-S2D & LERD & DGVI \\
\midrule
LSMOP & 17/36 & 12/17/7 & 0/34/2 & 5/29/2 & 10/20/6 & 4/30/2 & 5/28/3 & 4/31/1 \\
UF    & 34/40 & 1/31/8  & 0/40/0 & 0/40/0 & 4/34/2  & 0/40/0 & 0/36/4 & 0/38/2 \\
Overall & 51/76 & 13/48/15 & 0/74/2 & 5/69/2 & 14/54/8 & 4/70/2 & 5/64/7 & 4/69/3 \\
\bottomrule
\end{tabular}%
}
\endgroup
\end{table*}

\subsection{Ablation Study}

\label{subsec:ablation_study}
To further examine the contribution of each component in DDFS, five variants are compared with the full version of DDFS on the $500$-dimensional test problems. The detailed IGD results are provided in the supplementary material, and Table~\ref{tab:ablation_summary} summarizes the Wilcoxon statistical results of each variant against DDFS.

\begin{table}[!t]
	\centering
	\caption{Ablation study summary on 500-dimensional test problems.}
	\label{tab:ablation_summary}
	\small
	\setlength{\tabcolsep}{2.5pt}
	\renewcommand{\arraystretch}{1.05}
	\resizebox{\linewidth}{!}{%
		\begin{tabular}{lcc}
			\toprule
			Variant & Main removed or modified component & $+/-/=$ \\
			\midrule
			NoSensitivityRefinement & No split between $I_{\mathrm{ch}}$ and $I_{\mathrm{cl}}$ & 1/14/4 \\
			UniformFuzzy & Uniform fuzzy granularity for all variables & 0/13/6 \\
			FixedStage & Fixed FE-based stage transition & 1/15/3 \\
			NoStage3Disabling & Stage 3 keeps Stage-2 fuzzy updating & 1/17/1 \\
			RandomVariableCategories & Random assignment of variable categories & 0/13/6 \\
			\bottomrule
		\end{tabular}%
	}
\end{table}

In the NoSensitivityRefinement variant, convergence-related variables are no longer divided into high-sensitivity and low-sensitivity subsets. Specifically, diversity-related variables still use 50 and 500 grid partitions in Stages 1 and 2, respectively, while all convergence-related variables use 100 and 1000 grid partitions. This variant is worse than DDFS on 14 instances, showing that sensitivity refinement provides useful fine-grained guidance for protecting sensitive convergence directions while still allowing faster progress on less sensitive variables.

The UniformFuzzy variant keeps the same decision-variable analysis cost but applies the same fuzzy granularity to all variables, with 100 grid partitions in Stage 1 and 1000 grid partitions in Stage 2. It is outperformed by DDFS on 13 instances and never significantly outperforms DDFS. This result demonstrates that the performance gain does not simply come from applying fuzzy discretization, but from matching fuzzy granularities with the roles of different variables.

The FixedStage variant replaces the dual-indicator stage-transition mechanism with fixed FE-based switching at one-third and two-thirds of the total function evaluations. Its worse performance on 15 instances suggests that a time-driven schedule can be less reliable than state-driven stage control, because different problems may have different convergence and search-region contraction rhythms. The NoStage3Disabling variant continues to use the Stage-2 fuzzy granularity in Stage 3 and is worse than DDFS on 17 instances, suggesting that disabling fuzzy updating in the late stage is helpful for reducing unnecessary discretization interference. Finally, the RandomVariableCategories variant randomly assigns variables into three categories while preserving the same analysis cost, and it is worse than DDFS on 13 instances. This result suggests that the detected variable categories contain useful structural information rather than acting as arbitrary labels. These results further support the usefulness of establishing an explicit mapping between decision-variable roles and fuzzy search granularities.

\subsection{Sensitivity Analysis}

\label{subsec:sensitivity_analysis}
To further examine the robustness of DDFS with respect to key parameters, sensitivity analyses are conducted on the $500$-dimensional test problems. Two groups of parameters are considered, including the stage-transition parameters and the fuzzy-granularity parameters. The detailed IGD results are provided in the supplementary material, and Table~\ref{tab:parameter_sensitivity_summary} summarizes the Wilcoxon statistical results of different parameter settings against the default DDFS.

\begin{table}[!t]
\centering
\caption{Sensitivity analysis summary on 500-dimensional test problems.}
\label{tab:parameter_sensitivity_summary}
\small
\setlength{\tabcolsep}{3.0pt}
\renewcommand{\arraystretch}{1.05}
\resizebox{\linewidth}{!}{%
\begin{tabular}{llcc}
\toprule
Parameter group & Parameter & Setting & $+/-/=$ \\
\midrule
\multirow{6}{*}{Stage transition}
& $K$ & $2$ & 2/12/5 \\
& $K$ & $3$ & 1/8/10 \\
& $\tau_1$ & $0.5$ & 0/9/10 \\
& $\tau_1$ & $0.7$ & 2/10/7 \\
& $\tau_2$ & $0.2$ & 1/13/5 \\
& $\tau_2$ & $0.3$ & 2/9/8 \\
\midrule
\multirow{8}{*}{Fuzzy granularity}
& Diversity-related variables & $5/50$ & 2/2/15 \\
& Diversity-related variables & $20/200$ & 1/0/18 \\
& Diversity-related variables & $50/500$ & 1/0/18 \\
& High-sensitivity convergence-related variables & $500/5000$ & 0/0/19 \\
& High-sensitivity convergence-related variables & $2000/20000$ & 0/0/19 \\
& Low-sensitivity convergence-related variables & $50/500$ & 0/1/18 \\
& Low-sensitivity convergence-related variables & $200/2000$ & 0/2/17 \\
& Low-sensitivity convergence-related variables & $500/5000$ & 2/0/17 \\
\bottomrule
\end{tabular}%
}
\end{table}

As shown in Table~\ref{tab:parameter_sensitivity_summary}, the stage-transition parameters affect the stability of DDFS. When the hysteresis length is too small, such as $K=2$, the corresponding setting is worse than the default DDFS on 12 instances, indicating that overly frequent stage switching may make fuzzy updating sensitive to temporary fluctuations in the search state. Increasing $K$ improves the overall stability, and $K=4$ provides a more balanced result. The transition thresholds also influence performance. A small $\tau_2$, such as $\tau_2=0.2$, is worse than DDFS on 13 instances, suggesting that delaying the transition to the late stage may keep fuzzy updating active for too long and interfere with refinement. Overall, the default setting with $\tau_1=0.6$, $\tau_2=0.4$, and $K=4$ achieves a relatively stable balance between search-space compression and late-stage refinement.

For the fuzzy-granularity parameters, most settings are statistically similar to the default DDFS on the majority of test instances. In particular, the two settings for high-sensitivity convergence-related variables obtain 19 ties, showing that DDFS is not sensitive to moderate changes in the fine-grained updating scale. The results also indicate that the effectiveness of DDFS does not rely on a single exact grid number. Instead, the relative granularity hierarchy among different variable categories is more important. Diversity-related variables can tolerate relatively coarse grids to support broad exploration, high-sensitivity convergence-related variables require finer grids to avoid excessive disturbance, and low-sensitivity convergence-related variables are suitable for an intermediate scale. These results indicate that the proposed differentiated fuzzy search is relatively stable under a reasonable range of parameter settings.

\section{Conclusion}

This paper proposed DDFS, a decision variable analysis-guided differentiated fuzzy search method for large-scale multi-objective optimization. DDFS first classifies decision variables through single-variable perturbation analysis and further refines convergence-related variables according to their objective-response sensitivities. Based on the obtained variable categories, different fuzzy granularities are assigned to diversity-related variables, low-sensitivity convergence-related variables, and high-sensitivity convergence-related variables during offspring generation. In addition, a dual-indicator stage transition mechanism is designed to adaptively regulate the fuzzy updating intensity according to convergence improvement and decision-space dispersion. Experimental results on the LSMOP and UF benchmark suites show that DDFS generally achieves competitive performance against representative large-scale multi-objective evolutionary algorithms. The ablation and sensitivity studies further support the usefulness of variable-wise differentiated fuzzy granularities, sensitivity refinement, the dual-indicator stage transition mechanism, and late-stage disabling of fuzzy updating.

Although DDFS is implemented as a lightweight search module in this study, its empirical evaluation is mainly based on the LMOCSO instantiation. In addition, its variable categories are obtained once during initialization and remain fixed throughout evolution. This design avoids repeated variable-analysis costs, but may not fully capture possible changes in variable importance during the search process. Future work will investigate lightweight dynamic variable reanalysis strategies and examine the integration of differentiated fuzzy search with different base MOEAs, while maintaining the low-overhead nature of the proposed method.

\section*{Acknowledgment}
This research is supported by the National Natural Science Foundation of China (Nos. 62572419 and 62176228).

\bibliography{references}

\begin{thebibliography}{10}
\providecommand{\url}[1]{#1}
\csname url@samestyle\endcsname
\providecommand{\newblock}{\relax}
\providecommand{\bibinfo}[2]{#2}
\providecommand{\BIBentrySTDinterwordspacing}{\spaceskip=0pt\relax}
\providecommand{\BIBentryALTinterwordstretchfactor}{4}
\providecommand{\BIBentryALTinterwordspacing}{\spaceskip=\fontdimen2\font plus
\BIBentryALTinterwordstretchfactor\fontdimen3\font minus
  \fontdimen4\font\relax}
\providecommand{\BIBforeignlanguage}[2]{{%
\expandafter\ifx\csname l@#1\endcsname\relax
\typeout{** WARNING: IEEEtran.bst: No hyphenation pattern has been}%
\typeout{** loaded for the language `#1'. Using the pattern for}%
\typeout{** the default language instead.}%
\else
\language=\csname l@#1\endcsname
\fi
#2}}
\providecommand{\BIBdecl}{\relax}
\BIBdecl

\bibitem{pereira2022review}
J.~L.~J. Pereira, G.~A. Oliver, M.~B. Francisco, S.~S. Cunha~Jr, and G.~F.
  Gomes, ``A review of multi-objective optimization: methods and algorithms in
  mechanical engineering problems,'' \emph{Archives of Computational Methods in
  Engineering}, vol.~29, no.~4, pp. 2285--2308, 2022.

\bibitem{chen2023balancing}
L.~Chen, Y.~Xu, F.~Xu, Q.~Hu, and Z.~Tang, ``Balancing the trade-off between
  cost and reliability for wireless sensor networks: a multi-objective
  optimized deployment method,'' \emph{Applied Intelligence}, vol.~53, no.~8,
  pp. 9148--9173, 2023.

\bibitem{guo2026scalable}
W.~Guo, Y.~Xiao, Z.~Zhang, W.~Li, L.~Zhang, L.~Li, D.~Li, C.~Rohmann, H.~K.
  Bachem, and M.~Hinz, ``Scalable multi-objective optimization for robust
  traffic signal control in uncertain environments based on hierarchical
  reinforcement learning,'' \emph{IEEE Transactions on Intelligent
  Transportation Systems}, 2026.

\bibitem{zitzler1999spea}
E.~Zitzler and L.~Thiele, ``Multiobjective evolutionary algorithms: A
  comparative case study and the strength pareto approach,'' \emph{IEEE
  Transactions on Evolutionary Computation}, vol.~3, no.~4, pp. 257--271, 1999.

\bibitem{knowles1999paes}
J.~Knowles and D.~Corne, ``The pareto archived evolution strategy: A new
  baseline algorithm for pareto multiobjective optimisation,'' in
  \emph{Proceedings of the 1999 Congress on Evolutionary Computation}, 1999,
  pp. 98--105.

\bibitem{zitzler2004ibea}
E.~Zitzler and S.~K{\"u}nzli, ``Indicator-based selection in multiobjective
  search,'' in \emph{Parallel Problem Solving from Nature -- PPSN VIII}.\hskip
  1em plus 0.5em minus 0.4em\relax Springer, 2004, pp. 832--842.

\bibitem{beume2007smsemoa}
N.~Beume, B.~Naujoks, and M.~Emmerich, ``{SMS-EMOA}: Multiobjective selection
  based on dominated hypervolume,'' \emph{European Journal of Operational
  Research}, vol. 181, no.~3, pp. 1653--1669, 2007.

\bibitem{zhou2011survey}
A.~Zhou, B.-Y. Qu, H.~Li, S.-Z. Zhao, P.~N. Suganthan, and Q.~Zhang,
  ``Multiobjective evolutionary algorithms: A survey of the state of the art,''
  \emph{Swarm and Evolutionary Computation}, vol.~1, no.~1, pp. 32--49, 2011.

\bibitem{LSMOP}
R.~Cheng, Y.~Jin, M.~Olhofer \emph{et~al.}, ``Test problems for large-scale
  multi-objective and many-objective optimization,'' \emph{IEEE Transactions on
  Cybernetics}, vol.~47, no.~12, pp. 4108--4121, 2016.

\bibitem{LMF}
S.~Liu, Q.~Lin, K.-C. Wong, Q.~Li, and K.~C. Tan, ``Evolutionary large-scale
  multi-objective optimization: Benchmarks and algorithms,'' \emph{IEEE
  Transactions on Evolutionary Computation}, vol.~27, no.~3, pp. 401--415,
  2021.

\bibitem{tian2021evolutionary}
Y.~Tian, L.~Si, X.~Zhang, R.~Cheng, C.~He, K.~C. Tan, and Y.~Jin,
  ``Evolutionary large-scale multi-objective optimization: A survey,''
  \emph{ACM Computing Surveys (CSUR)}, vol.~54, no.~8, pp. 1--34, 2021.

\bibitem{liu2023survey}
S.~Liu, Q.~Lin, J.~Li, and K.~C. Tan, ``A survey on learnable evolutionary
  algorithms for scalable multiobjective optimization,'' \emph{IEEE
  Transactions on Evolutionary Computation}, vol.~27, no.~6, pp. 1941--1961,
  2023.

\bibitem{liu2024large}
J.~Liu, R.~Sarker, S.~Elsayed, D.~Essam, and N.~Siswanto, ``Large-scale
  evolutionary optimization: A review and comparative study,'' \emph{Swarm and
  Evolutionary Computation}, vol.~85, p. 101466, 2024.

\bibitem{CCGDE3}
L.~M. Antonio and C.~A.~C. Coello, ``Use of cooperative coevolution for solving
  large scale multi-objective optimization problems,'' in \emph{2013 IEEE
  Congress on Evolutionary Computation}.\hskip 1em plus 0.5em minus 0.4em\relax
  IEEE, 2013, pp. 2758--2765.

\bibitem{miguel2016decomposition}
L.~Miguel~Antonio and C.~A. Coello~Coello, ``Decomposition-based approach for
  solving large scale multi-objective problems,'' in \emph{Parallel Problem
  Solving from Nature--PPSN XIV: 14th International Conference, Edinburgh, UK,
  September 17-21, 2016, Proceedings 14}.\hskip 1em plus 0.5em minus
  0.4em\relax Springer, 2016, pp. 525--534.

\bibitem{CC2}
P.~Xu, W.~Luo, X.~Lin, J.~Zhang, Y.~Qiao, and X.~Wang, ``Constraint-objective
  cooperative coevolution for large-scale constrained optimization,'' \emph{ACM
  Transactions on Evolutionary Learning and Optimization}, vol.~1, no.~3, pp.
  1--26, 2021.

\bibitem{WOF}
H.~Zille, H.~Ishibuchi, S.~Mostaghim, and Y.~Nojima, ``A framework for
  large-scale multiobjective optimization based on problem transformation,''
  \emph{IEEE Transactions on Evolutionary Computation}, vol.~22, no.~2, pp.
  260--275, 2018.

\bibitem{ordered}
------, ``A framework for large-scale multi-objective optimization based on
  problem transformation,'' \emph{IEEE Transactions on Evolutionary
  Computation}, vol.~22, no.~2, pp. 260--275, 2017.

\bibitem{LSMOF}
C.~He, L.~Li, Y.~Tian, X.~Zhang, R.~Cheng, Y.~Jin, and X.~Yao, ``Accelerating
  large-scale multi-objective optimization via problem reformulation,''
  \emph{IEEE Transactions on Evolutionary Computation}, vol.~23, no.~6, pp.
  949--961, 2019.

\bibitem{LSMOEA-RD}
Z.~Xiong, X.~Wang, Y.~Li, W.~Feng, and Y.~Liu, ``A problem transformation-based
  and decomposition-based evolutionary algorithm for large-scale multiobjective
  optimization,'' \emph{Applied Soft Computing}, vol. 150, p. 111081, 2024.

\bibitem{CSO}
R.~Cheng and Y.~Jin, ``A competitive swarm optimizer for large scale
  optimization,'' \emph{IEEE transactions on cybernetics}, vol.~45, no.~2, pp.
  191--204, 2014.

\bibitem{LMOCSO}
Y.~Tian, X.~Zheng, X.~Zhang, and Y.~Jin, ``Efficient large-scale
  multi-objective optimization based on a competitive swarm optimizer,''
  \emph{IEEE Transactions on Cybernetics}, vol.~50, no.~8, pp. 3696--3708,
  2019.

\bibitem{E-CSO}
\BIBentryALTinterwordspacing
S.~Qi, J.~Zou, S.~Yang, Y.~Jin, J.~Zheng, and X.~Yang, ``A self-exploratory
  competitive swarm optimization algorithm for large-scale multiobjective
  optimization,'' \emph{Information Sciences}, vol. 609, pp. 1601--1620, 2022.
  [Online]. Available:
  \url{https://www.sciencedirect.com/science/article/pii/S0020025522007988}
\BIBentrySTDinterwordspacing

\bibitem{LMOEA-S2D}
J.~Zou, L.~Tang, Y.~Liu, S.~Yang, and S.~Wang, ``A two-stage direction-guided
  evolutionary algorithm for large-scale multiobjective optimization,''
  \emph{Information Sciences}, vol. 674, p. 120719, 2024.

\bibitem{LM-DAS}
J.~Zhang, L.~Wei, R.~Fan, H.~Sun, and Z.~Hu, ``Solve large-scale many-objective
  optimization problems based on dual analysis of objective space and decision
  space,'' \emph{Swarm and Evolutionary Computation}, vol.~70, p. 101045, 2022.

\bibitem{DGEA}
C.~He, R.~Cheng, and D.~Yazdani, ``Adaptive offspring generation for
  evolutionary large-scale multiobjective optimization,'' \emph{IEEE
  Transactions on Systems, Man, and Cybernetics: Systems}, vol.~52, no.~2, pp.
  786--798, 2020.

\bibitem{FDV}
X.~Yang, J.~Zou, S.~Yang, J.~Zheng, and Y.~Liu, ``A fuzzy decision variables
  framework for large-scale multiobjective optimization,'' \emph{IEEE
  Transactions on Evolutionary Computation}, vol.~27, no.~3, pp. 445--459,
  2021.

\bibitem{FDVDS2023}
S.-T. Wang, J.-H. Zheng, J.~Zou, Y.~Liu, S.-X. Yang, and Y.-J. Zou, ``A fuzzy
  decision variables framework based on directed sampling for large-scale
  multiobjective optimization,'' in \emph{Proceedings of the Companion
  Conference on Genetic and Evolutionary Computation}.\hskip 1em plus 0.5em
  minus 0.4em\relax ACM, 2023, pp. 419--422.

\bibitem{MOEA/DVA}
X.~Ma, F.~Liu, Y.~Qi, X.~Wang, L.~Li, L.~Jiao, M.~Yin, and M.~Gong, ``A
  multi-objective evolutionary algorithm based on decision variable analyses
  for multi-objective optimization problems with large-scale variables,''
  \emph{IEEE Transactions on Evolutionary Computation}, vol.~20, no.~2, pp.
  275--298, 2015.

\bibitem{LMEA}
X.~Zhang, Y.~Tian, R.~Cheng, and Y.~Jin, ``A decision variable clustering-based
  evolutionary algorithm for large-scale many-objective optimization,''
  \emph{IEEE Transactions on Evolutionary Computation}, vol.~22, no.~1, pp.
  97--112, 2016.

\bibitem{LERD}
C.~He, R.~Cheng, L.~Li, K.~C. Tan, and Y.~Jin, ``Large-scale multi-objective
  optimization via reformulated decision variable analysis,'' \emph{IEEE
  Transactions on Evolutionary Computation}, vol.~28, no.~1, pp. 47--61, 2024.

\bibitem{WOF-PAG}
M.~Zhang, W.~Li, L.~Zhang, H.~Jin, Y.~Mu, and L.~Wang, ``A pearson
  correlation-based adaptive variable grouping method for large-scale
  multi-objective optimization,'' \emph{Information Sciences}, vol. 639, p.
  118737, 2023.

\bibitem{deb2014nsgaIII}
K.~Deb and H.~Jain, ``An evolutionary many-objective optimization algorithm
  using reference-point-based nondominated sorting approach, part {I}: Solving
  problems with box constraints,'' \emph{IEEE Transactions on Evolutionary
  Computation}, vol.~18, no.~4, pp. 577--601, 2014.

\bibitem{cheng2016rvea}
R.~Cheng, Y.~Jin, M.~Olhofer, and B.~Sendhoff, ``A reference vector guided
  evolutionary algorithm for many-objective optimization,'' \emph{IEEE
  Transactions on Evolutionary Computation}, vol.~20, no.~5, pp. 773--791,
  2016.

\bibitem{DGVI2025}
Y.~Chen, Z.~Zhou, Y.~Liu, L.~Hu, Z.~Zou, L.~Xu, Z.~Gan, and C.~Ouyang,
  ``Dynamic grouping with a self-aware computational resource allocation for
  large-scale multi-objective optimization,'' \emph{IEEE Transactions on
  Evolutionary Computation}, 2025.

\bibitem{APT}
S.~Liu, J.~Li, Q.~Lin, Y.~Tian, J.~Li, and K.~C. Tan, ``Evolutionary
  large-scale multi-objective optimization via autoencoder-based problem
  transformation,'' \emph{IEEE Transactions on Emerging Topics in Computational
  Intelligence}, 2024.

\bibitem{IBde2026}
L.~Si, Z.~Wang, X.~Zhang, and Y.~Tian, ``Information bottleneck theory-guided
  dimension reduction for large-scale multi-objective optimization,''
  \emph{IEEE Transactions on Evolutionary Computation}, 2026.

\bibitem{UF2009}
Q.~Zhang, A.~Zhou, S.~Zhao, P.~N. Suganthan, W.~Liu, and S.~Tiwari,
  ``Multiobjective optimization test instances for the cec 2009 special session
  and competition,'' University of Essex and Nanyang Technological University,
  Tech. Rep., 2008.

\bibitem{IGD}
A.~Zhou, Y.~Jin, Q.~Zhang, B.~Sendhoff, and E.~Tsang, ``Combining model-based
  and genetics-based offspring generation for multi-objective optimization
  using a convergence criterion,'' in \emph{2006 IEEE international conference
  on evolutionary computation}.\hskip 1em plus 0.5em minus 0.4em\relax IEEE,
  2006, pp. 892--899.

\bibitem{HV}
E.~Zitzler, ``Multi-objective evolutionary algorithms: a comparative case study
  and the strength pareto approach,'' \emph{IEEE Transactions on Evolutionary
  Computaion}, vol.~2, no.~3, pp. 221--248, 1994.

\bibitem{FLEA}
L.~Li, C.~He, R.~Cheng, H.~Li, L.~Pan, and Y.~Jin, ``A fast sampling based
  evolutionary algorithm for million-dimensional multiobjective optimization,''
  \emph{Swarm and Evolutionary Computation}, vol.~75, p. 101181, 2022.

\bibitem{platemo}
Y.~Tian, R.~Cheng, X.~Zhang, and Y.~Jin, ``Platemo: A matlab platform for
  evolutionary multi-objective optimization [educational forum],'' \emph{IEEE
  Computational Intelligence Magazine}, vol.~12, no.~4, pp. 73--87, 2017.

\bibitem{Wilcoxon2013}
W.~Haynes, ``Wilcoxon rank sum test,'' in \emph{Encyclopedia of Systems
  Biology}.\hskip 1em plus 0.5em minus 0.4em\relax Springer, 2013, pp.
  2354--2355.

\end{thebibliography}

\clearpage
\setcounter{section}{0}
\setcounter{table}{0}
\setcounter{figure}{0}
\renewcommand{\thesection}{S\arabic{section}}
\renewcommand{\thetable}{S-\Roman{table}}
\renewcommand{\thefigure}{S-\arabic{figure}}
\markboth{SUPPLEMENTARY MATERIAL FOR DDFS}%
{Xiao et al.: Supplementary Material for Decision Variable Analysis-Guided Differentiated Fuzzy Search}
\twocolumn[
\begin{center}
{\LARGE\bfseries Supplementary Material for ``Decision Variable Analysis-Guided Differentiated Fuzzy Search for Large-Scale Multi-Objective Optimization''\par}
\vspace{1em}
{\large Boxi Xiao, Hui Bai, Jinhua Zheng, Yu Li, and Juan Zou\par}
\end{center}
\vspace{1em}
]
\section{Detailed HV Results of Comparative Experiments}
\label{supp:hv_results}

This section reports the detailed hypervolume (HV) results of the comparative experiments on the LSMOP and UF test problems with $300$, $500$, $800$, and $1000$ decision variables. For HV, a larger value indicates better performance. The symbols ``+'', ``-'', and ``='' indicate that the corresponding compared algorithm is significantly better than, significantly worse than, and statistically similar to DDFS, respectively. The best mean HV value for each test instance is highlighted with gray background and bold font.

Tables~\ref{tab:supp_hv_lsmop_all_dims} and~\ref{tab:supp_hv_uf_all_dims} report the HV results of the comparative experiments. Since HV evaluates the dominated objective-space volume, a larger value indicates better convergence and distribution with respect to the selected reference point. Overall, the HV results are broadly consistent with the IGD-based observations in the main manuscript, while also showing that several compared algorithms remain competitive on some test instances.

On the LSMOP suite, DDFS obtains the best mean HV values on $17$ out of $36$ instances. The best-HV results are mainly observed on LSMOP1, LSMOP4, LSMOP6, LSMOP8, and LSMOP9, suggesting that the proposed differentiated fuzzy search provides useful guidance for several large-scale problems. The statistical results also show that DDFS is significantly better than CCGDE3, FLEA, LMOEA-S2D, LERD, and DGVI on $34$, $29$, $30$, $28$, and $31$ instances, respectively. Meanwhile, FDV and LMOCSO obtain significantly better HV values than DDFS on $12$ and $10$ instances, respectively. These results indicate that DDFS remains competitive on the LSMOP suite, but its HV advantage varies across different problem structures.

On the UF suite, DDFS shows favorable HV behavior. It obtains the best mean HV values on $34$ out of $40$ instances and is significantly better than CCGDE3, FLEA, and LMOEA-S2D on all $40$ instances. DDFS is also significantly better than FDV, LMOCSO, LERD, and DGVI on $31$, $34$, $36$, and $38$ instances, respectively. The advantage is relatively consistent on UF1, UF4, UF6, UF7, UF8, and UF9, while LMOCSO, FDV, or LERD achieves better HV on a few cases such as UF2, UF3, UF5, and UF10. Therefore, the HV results provide complementary evidence that DDFS can obtain competitive solution sets on many UF problems, while the performance difference remains problem-dependent.

Combining the IGD results in the main manuscript and the HV results in this supplementary material, DDFS generally exhibits competitive performance in terms of both distance-based and volume-based indicators. The results also indicate that the proposed variable-wise fuzzy granularity and stage-adaptive fuzzy-updating control are helpful in many high-dimensional test instances, while leaving room for further improvement on problems where other search strategies achieve better HV values.

\clearpage
\onecolumn
\begin{landscape}

\begin{table}[H]
\vspace{-2mm}
\centering
\caption{Detailed HV results obtained by DDFS and the compared algorithms on LSMOP test problems with 300, 500, 800, and 1000 decision variables.}\vspace{-1mm}
\label{tab:supp_hv_lsmop_all_dims}
\tiny
\setlength{\tabcolsep}{1.0pt}
\renewcommand{\arraystretch}{0.72}
\resizebox{\linewidth}{!}{%
\begin{tabular}{ccc|c|c|c|c|c|c|c|c}
\toprule
Problem & M & D & FDV & CCGDE3 & FLEA & LMOCSO & LMOEA-S2D & LERD & DGVI & DDFS \\
\midrule
\multirow{4}{*}{LSMOP1} & \multirow{4}{*}{2} & 300 & 5.8323e-1 (2.09e-4) - & 0.0000e+0 (0.00e+0) - & 1.9284e-1 (5.72e-2) - & 5.8323e-1 (1.78e-4) - & 2.5385e-1 (1.22e-2) - & 5.8017e-1 (3.49e-4) - & 5.2964e-1 (1.61e-2) - & \best{5.8358e-1 (9.27e-5)} \\
 &  & 500 & 5.8369e-1 (9.81e-5) - & 0.0000e+0 (0.00e+0) - & 2.2422e-1 (4.69e-2) - & \best{5.8440e-1 (5.04e-5) +} & 2.5710e-1 (4.85e-3) - & 5.8080e-1 (2.00e-4) - & 5.5914e-1 (7.02e-3) - & 5.8408e-1 (4.05e-5) \\
 &  & 800 & 5.8373e-1 (1.21e-4) - & 0.0000e+0 (0.00e+0) - & 2.3543e-1 (2.78e-2) - & 5.8388e-1 (5.36e-5) - & 2.5572e-1 (2.32e-3) - & 5.7950e-1 (2.54e-4) - & 5.5857e-1 (7.32e-3) - & \best{5.8437e-1 (3.70e-5)} \\
 &  & 1000 & 5.8373e-1 (1.19e-4) - & 0.0000e+0 (0.00e+0) - & 2.3287e-1 (2.75e-2) - & 5.8345e-1 (6.84e-5) - & 2.5659e-1 (2.99e-3) - & 5.7886e-1 (2.52e-4) - & 5.5774e-1 (5.12e-3) - & \best{5.8444e-1 (6.62e-5)} \\
\midrule
\multirow{4}{*}{LSMOP2} & \multirow{4}{*}{2} & 300 & \best{5.8131e-1 (6.45e-4) +} & 4.5612e-1 (1.95e-3) - & 5.5205e-1 (3.33e-3) + & 5.3087e-1 (1.33e-3) - & 5.6389e-1 (3.44e-4) + & 5.5969e-1 (1.21e-2) + & 5.1675e-1 (2.75e-3) - & 5.4452e-1 (6.43e-3) \\
 &  & 500 & \best{5.8273e-1 (1.91e-4) +} & 4.9771e-1 (1.16e-3) - & 5.6583e-1 (2.79e-3) + & 5.5576e-1 (3.94e-4) - & 5.7080e-1 (2.06e-3) + & 5.7652e-1 (2.80e-3) + & 5.3659e-1 (2.49e-3) - & 5.5694e-1 (7.54e-4) \\
 &  & 800 & \best{5.8330e-1 (9.20e-5) +} & 5.2495e-1 (1.16e-3) - & 5.7200e-1 (1.12e-3) + & 5.6540e-1 (2.12e-4) = & 5.7468e-1 (1.33e-3) + & 5.7669e-1 (2.81e-3) + & 5.4736e-1 (1.18e-3) - & 5.6538e-1 (3.89e-4) \\
 &  & 1000 & \best{5.8347e-1 (9.21e-5) +} & 5.3558e-1 (7.98e-4) - & 5.7489e-1 (1.14e-3) + & 5.6930e-1 (1.93e-4) + & 5.7723e-1 (4.70e-4) + & 5.7746e-1 (1.25e-3) + & 5.5325e-1 (5.62e-4) - & 5.6791e-1 (3.19e-4) \\
\midrule
\multirow{4}{*}{LSMOP3} & \multirow{4}{*}{2} & 300 & 9.0744e-2 (3.48e-5) + & 0.0000e+0 (0.00e+0) - & 0.0000e+0 (0.00e+0) - & \best{9.0889e-2 (6.61e-6) +} & 0.0000e+0 (0.00e+0) - & 9.0734e-2 (6.78e-5) + & 0.0000e+0 (0.00e+0) - & 6.3962e-2 (3.66e-2) \\
 &  & 500 & 9.0838e-2 (1.34e-5) - & 0.0000e+0 (0.00e+0) - & 0.0000e+0 (0.00e+0) - & 9.0907e-2 (2.69e-6) - & 0.0000e+0 (0.00e+0) - & 8.4832e-2 (2.35e-2) - & 0.0000e+0 (0.00e+0) - & \best{9.2957e-2 (1.52e-2)} \\
 &  & 800 & 9.0855e-2 (7.13e-6) + & 0.0000e+0 (0.00e+0) - & 0.0000e+0 (0.00e+0) - & \best{9.0906e-2 (3.23e-7) +} & 0.0000e+0 (0.00e+0) - & 7.8770e-2 (3.20e-2) - & 0.0000e+0 (0.00e+0) - & 9.0223e-2 (1.76e-4) \\
 &  & 1000 & 9.0860e-2 (6.22e-6) + & 0.0000e+0 (0.00e+0) - & 0.0000e+0 (0.00e+0) - & \best{9.0907e-2 (3.03e-7) +} & 0.0000e+0 (0.00e+0) - & 7.8768e-2 (3.20e-2) - & 0.0000e+0 (0.00e+0) - & 9.0303e-2 (1.14e-4) \\
\midrule
\multirow{4}{*}{LSMOP4} & \multirow{4}{*}{2} & 300 & 5.8121e-1 (1.55e-3) - & 4.0325e-1 (7.30e-3) - & 4.9555e-1 (5.83e-3) - & 5.7074e-1 (3.32e-4) - & 5.0859e-1 (1.35e-3) - & 5.7217e-1 (7.44e-4) - & 5.4966e-1 (3.89e-3) - & \best{5.8301e-1 (2.43e-4)} \\
 &  & 500 & 5.8321e-1 (3.80e-4) - & 4.5466e-1 (2.81e-3) - & 5.2650e-1 (5.78e-3) - & 5.8346e-1 (5.23e-4) - & 5.3616e-1 (1.21e-3) - & 5.7589e-1 (9.13e-4) - & 5.6889e-1 (2.25e-3) - & \best{5.8385e-1 (5.93e-5)} \\
 &  & 800 & 5.8397e-1 (1.10e-4) = & 4.9110e-1 (1.46e-3) - & 5.5022e-1 (2.90e-3) - & 5.8325e-1 (3.40e-4) - & 5.5335e-1 (9.09e-4) - & 5.7796e-1 (3.60e-4) - & 5.6491e-1 (5.31e-3) - & \best{5.8403e-1 (4.87e-5)} \\
 &  & 1000 & 5.8408e-1 (1.34e-4) = & 5.0464e-1 (1.48e-3) - & 5.5786e-1 (1.43e-3) - & \best{5.8416e-1 (2.36e-5) =} & 5.5984e-1 (4.66e-4) - & 5.7814e-1 (9.64e-4) - & 5.6494e-1 (6.41e-3) - & 5.8413e-1 (1.45e-4) \\
\midrule
\multirow{4}{*}{LSMOP5} & \multirow{4}{*}{2} & 300 & \best{3.4626e-1 (2.53e-4) +} & 0.0000e+0 (0.00e+0) - & 9.0909e-2 (2.87e-17) - & 3.4612e-1 (2.63e-4) + & 9.0909e-2 (1.41e-8) - & 2.9562e-1 (7.63e-2) - & 3.3843e-1 (1.52e-3) - & 3.4527e-1 (2.51e-4) \\
 &  & 500 & 3.4690e-1 (2.03e-4) + & 0.0000e+0 (0.00e+0) - & 9.0909e-2 (2.87e-17) - & \best{3.4813e-1 (1.05e-4) +} & 9.0909e-2 (4.97e-8) - & 3.3403e-1 (4.44e-2) - & 3.4406e-1 (3.04e-4) - & 3.4659e-1 (1.07e-4) \\
 &  & 800 & 3.4668e-1 (1.82e-4) + & 0.0000e+0 (0.00e+0) - & 9.0909e-2 (2.87e-17) - & \best{3.4789e-1 (1.02e-4) +} & 9.0909e-2 (3.29e-8) - & 3.4368e-1 (3.82e-4) - & 3.4459e-1 (5.26e-4) - & 3.4562e-1 (2.07e-4) \\
 &  & 1000 & 3.4657e-1 (1.44e-4) + & 0.0000e+0 (0.00e+0) - & 9.0909e-2 (2.87e-17) - & \best{3.4810e-1 (8.61e-5) +} & 9.0909e-2 (3.85e-8) - & 3.4283e-1 (4.91e-4) - & 3.4497e-1 (2.03e-4) + & 3.4436e-1 (2.75e-4) \\
\midrule
\multirow{4}{*}{LSMOP6} & \multirow{4}{*}{2} & 300 & 1.0046e-1 (2.49e-2) + & 0.0000e+0 (0.00e+0) - & \best{1.3940e-1 (4.07e-3) +} & 8.8608e-2 (2.08e-4) - & 7.4747e-2 (2.41e-2) - & 8.9209e-2 (4.34e-4) - & 8.3294e-2 (2.58e-3) - & 9.5377e-2 (6.55e-2) \\
 &  & 500 & 8.6086e-2 (9.14e-3) - & 0.0000e+0 (0.00e+0) - & 1.3720e-1 (5.34e-2) - & 8.9823e-2 (9.22e-5) - & 8.9483e-2 (9.99e-3) - & 9.5275e-2 (1.90e-2) - & 8.4244e-2 (1.77e-3) - & \best{2.0573e-1 (5.51e-2)} \\
 &  & 800 & 9.7370e-2 (2.52e-2) - & 1.4284e-2 (1.98e-3) - & 1.7086e-1 (1.00e-3) - & 8.9598e-2 (1.13e-4) - & 1.1056e-1 (2.37e-2) - & 7.8236e-2 (3.18e-2) - & 7.6125e-2 (2.27e-3) - & \best{2.6166e-1 (2.42e-2)} \\
 &  & 1000 & 8.6576e-2 (7.72e-4) - & 2.9109e-2 (1.49e-3) - & 1.6512e-1 (2.41e-2) - & 8.9674e-2 (9.87e-5) - & 1.1218e-1 (2.57e-2) - & 9.4674e-2 (1.69e-2) - & 6.9624e-2 (4.53e-3) - & \best{2.7594e-1 (1.35e-2)} \\
\midrule
\multirow{4}{*}{LSMOP7} & \multirow{4}{*}{2} & 300 & 0.0000e+0 (0.00e+0) = & 0.0000e+0 (0.00e+0) = & 0.0000e+0 (0.00e+0) = & \best{2.3752e-2 (4.97e-2) +} & 0.0000e+0 (0.00e+0) = & 0.0000e+0 (0.00e+0) = & 8.7272e-4 (3.38e-3) = & 0.0000e+0 (0.00e+0) \\
 &  & 500 & 2.7711e-3 (8.36e-3) = & 0.0000e+0 (0.00e+0) - & 0.0000e+0 (0.00e+0) - & 2.0863e-2 (5.59e-2) = & 0.0000e+0 (0.00e+0) - & 0.0000e+0 (0.00e+0) - & \best{8.0464e-2 (3.30e-2) +} & 7.5010e-2 (1.17e-1) \\
 &  & 800 & 1.5912e-3 (6.16e-3) = & 0.0000e+0 (0.00e+0) - & 0.0000e+0 (0.00e+0) - & 5.2034e-2 (1.08e-1) = & 0.0000e+0 (0.00e+0) - & 4.7172e-3 (1.83e-2) = & \best{9.9158e-2 (2.90e-2) +} & 5.0227e-2 (8.43e-2) \\
 &  & 1000 & 3.2787e-3 (9.76e-3) = & 0.0000e+0 (0.00e+0) = & 0.0000e+0 (0.00e+0) = & 5.4303e-2 (1.12e-1) = & 0.0000e+0 (0.00e+0) = & 0.0000e+0 (0.00e+0) = & \best{1.0167e-1 (3.59e-2) +} & 8.5080e-3 (3.30e-2) \\
\midrule
\multirow{4}{*}{LSMOP8} & \multirow{4}{*}{2} & 300 & 3.4005e-1 (2.10e-3) - & 0.0000e+0 (0.00e+0) - & 9.0909e-2 (2.87e-17) - & 3.1527e-1 (4.62e-3) - & 9.0909e-2 (2.87e-17) - & 3.2282e-1 (4.68e-3) - & 3.1527e-1 (3.58e-3) - & \best{3.4601e-1 (8.47e-4)} \\
 &  & 500 & 3.4316e-1 (2.76e-3) - & 0.0000e+0 (0.00e+0) - & 9.0909e-2 (2.87e-17) - & 3.3945e-1 (4.20e-3) - & 9.0909e-2 (3.87e-8) - & 3.3330e-1 (3.71e-3) - & 3.3634e-1 (2.12e-3) - & \best{3.4743e-1 (1.07e-4)} \\
 &  & 800 & 3.4488e-1 (1.79e-3) - & 0.0000e+0 (0.00e+0) - & 9.0909e-2 (2.87e-17) - & 3.4181e-1 (2.20e-3) - & 9.0909e-2 (2.87e-17) - & 3.3647e-1 (1.63e-3) - & 3.4088e-1 (8.17e-4) - & \best{3.4690e-1 (1.79e-4)} \\
 &  & 1000 & 3.4519e-1 (1.62e-3) = & 0.0000e+0 (0.00e+0) - & 9.0909e-2 (2.87e-17) - & \best{3.4595e-1 (2.54e-4) =} & 9.0909e-2 (4.26e-8) - & 3.3726e-1 (1.30e-3) - & 3.4197e-1 (9.03e-4) - & 3.4581e-1 (9.27e-4) \\
\midrule
\multirow{4}{*}{LSMOP9} & \multirow{4}{*}{2} & 300 & 1.7313e-1 (5.94e-2) - & 0.0000e+0 (0.00e+0) - & 1.8663e-1 (2.65e-2) - & 1.9138e-1 (5.71e-3) - & 1.5363e-1 (4.31e-2) - & 1.2514e-1 (3.81e-2) - & 2.2370e-1 (5.06e-3) - & \best{2.3689e-1 (1.76e-2)} \\
 &  & 500 & 2.4158e-1 (9.43e-5) - & 0.0000e+0 (0.00e+0) - & 2.1255e-1 (2.95e-5) - & 2.4223e-1 (4.17e-5) - & 1.8489e-1 (2.79e-2) - & 1.7480e-1 (1.37e-4) - & 2.4050e-1 (3.59e-4) - & \best{2.4237e-1 (5.46e-5)} \\
 &  & 800 & 2.4130e-1 (1.08e-4) - & 0.0000e+0 (0.00e+0) - & 2.2309e-1 (1.77e-5) - & 2.4053e-1 (1.64e-3) - & 1.7027e-1 (2.49e-2) - & 1.7439e-1 (1.37e-4) - & 2.4119e-1 (2.59e-4) - & \best{2.4263e-1 (2.84e-5)} \\
 &  & 1000 & 2.4073e-1 (2.39e-4) - & 0.0000e+0 (0.00e+0) - & 2.2644e-1 (1.24e-5) - & 2.4241e-1 (9.56e-5) - & 1.8156e-1 (2.68e-2) - & 1.7367e-1 (2.55e-4) - & 2.4109e-1 (3.58e-4) - & \best{2.4272e-1 (2.04e-5)} \\
\midrule
+/-/= &  &  & 12/17/7 & 0/34/2 & 5/29/2 & 10/20/6 & 4/30/2 & 5/28/3 & 4/31/1 &  \\
\bottomrule
\end{tabular}%
}
\end{table}
\end{landscape}

\clearpage
\begin{landscape}

\begin{table}[H]
\vspace{-2mm}
\centering
\caption{Detailed HV results obtained by DDFS and the compared algorithms on UF test problems with 300, 500, 800, and 1000 decision variables.}\vspace{-1mm}
\label{tab:supp_hv_uf_all_dims}
\tiny
\setlength{\tabcolsep}{1.0pt}
\renewcommand{\arraystretch}{0.72}
\resizebox{\linewidth}{!}{%
\begin{tabular}{ccc|c|c|c|c|c|c|c|c}
\toprule
Problem & M & D & FDV & CCGDE3 & FLEA & LMOCSO & LMOEA-S2D & LERD & DGVI & DDFS \\
\midrule
\multirow{4}{*}{UF1} & \multirow{4}{*}{2} & 300 & 5.2580e-1 (2.52e-1) - & 3.1514e-2 (2.02e-2) - & 5.2351e-2 (7.71e-2) - & 6.0753e-1 (6.94e-2) - & 3.9133e-1 (4.65e-3) - & 6.1185e-1 (2.90e-2) - & 6.0188e-1 (1.39e-2) - & \best{7.1710e-1 (2.29e-4)} \\
 &  & 500 & 5.2858e-1 (2.50e-1) - & 1.8789e-2 (1.22e-2) - & 4.1735e-2 (5.03e-2) - & 7.0188e-1 (4.33e-2) - & 3.9988e-1 (7.18e-3) - & 6.5779e-1 (4.09e-2) - & 6.3579e-1 (4.52e-3) - & \best{7.1967e-1 (8.00e-5)} \\
 &  & 800 & 6.2209e-1 (2.02e-1) = & 9.5261e-3 (8.67e-3) - & 4.2349e-2 (5.84e-2) - & 6.9885e-1 (4.38e-2) - & 3.8839e-1 (8.07e-3) - & 6.4269e-1 (4.62e-2) - & 6.3526e-1 (1.21e-2) - & \best{7.1937e-1 (1.13e-4)} \\
 &  & 1000 & 5.5053e-1 (2.62e-1) - & 7.7970e-3 (6.47e-3) - & 2.7883e-2 (3.66e-2) - & 7.1865e-1 (1.64e-4) - & 3.8505e-1 (5.10e-3) - & 6.2839e-1 (4.36e-2) - & 6.2805e-1 (1.40e-2) - & \best{7.2028e-1 (1.39e-3)} \\
\midrule
\multirow{4}{*}{UF2} & \multirow{4}{*}{2} & 300 & 7.1155e-1 (1.32e-3) - & 3.3725e-1 (4.87e-2) - & 5.5014e-1 (1.30e-2) - & 7.0371e-1 (1.37e-3) - & 6.1983e-1 (3.30e-3) - & 6.9064e-1 (1.80e-3) - & 6.6511e-1 (2.43e-3) - & \best{7.1286e-1 (1.32e-3)} \\
 &  & 500 & 7.1545e-1 (4.30e-4) - & 3.3767e-1 (3.71e-2) - & 5.4873e-1 (1.51e-2) - & 7.1711e-1 (8.04e-4) - & 6.2036e-1 (1.31e-3) - & 6.9691e-1 (1.49e-3) - & 6.7907e-1 (2.48e-3) - & \best{7.1833e-1 (4.97e-4)} \\
 &  & 800 & 7.1649e-1 (4.98e-4) = & 3.0370e-1 (4.74e-2) - & 5.5218e-1 (1.01e-2) - & \best{7.1798e-1 (6.73e-4) +} & 6.2035e-1 (1.10e-3) - & 6.9335e-1 (1.33e-3) - & 6.7495e-1 (2.83e-3) - & 7.1634e-1 (8.24e-4) \\
 &  & 1000 & 7.1699e-1 (3.69e-4) + & 3.0091e-1 (4.33e-2) - & 5.5928e-1 (8.81e-3) - & \best{7.1906e-1 (3.21e-4) +} & 6.2085e-1 (1.14e-3) - & 6.8914e-1 (1.37e-3) - & 6.6654e-1 (2.92e-3) - & 7.1541e-1 (6.11e-4) \\
\midrule
\multirow{4}{*}{UF3} & \multirow{4}{*}{2} & 300 & 7.1540e-1 (4.39e-3) - & 1.8480e-1 (3.54e-2) - & 5.4903e-1 (4.17e-3) - & 7.1548e-1 (6.99e-4) - & 5.6279e-1 (3.36e-3) - & 7.1220e-1 (1.64e-3) - & 6.7650e-1 (5.08e-3) - & \best{7.1664e-1 (3.99e-4)} \\
 &  & 500 & 7.1901e-1 (2.06e-3) - & 1.8687e-1 (2.24e-2) - & 5.6412e-1 (3.01e-3) - & 7.1973e-1 (1.14e-4) - & 5.7292e-1 (2.76e-3) - & 7.1840e-1 (4.82e-4) - & 6.9938e-1 (6.10e-3) - & \best{7.2103e-1 (7.11e-5)} \\
 &  & 800 & 7.1864e-1 (2.29e-3) - & 1.6337e-1 (3.68e-2) - & 5.7383e-1 (3.11e-3) - & 7.1975e-1 (2.09e-4) - & 5.8096e-1 (1.38e-3) - & 7.1824e-1 (1.79e-4) - & 7.0269e-1 (4.57e-3) - & \best{7.2095e-1 (1.24e-4)} \\
 &  & 1000 & 7.1966e-1 (1.92e-3) = & 1.4989e-1 (3.74e-2) - & 5.7817e-1 (2.73e-3) - & \best{7.2133e-1 (9.80e-5) +} & 5.8368e-1 (1.19e-3) - & 7.1725e-1 (4.30e-4) - & 7.0477e-1 (3.74e-3) - & 7.1927e-1 (2.05e-4) \\
\midrule
\multirow{4}{*}{UF4} & \multirow{4}{*}{2} & 300 & 3.2542e-1 (7.08e-3) - & 2.6062e-1 (5.08e-3) - & 2.8170e-1 (2.73e-3) - & 2.8699e-1 (3.42e-3) - & 3.5941e-1 (3.88e-4) - & 3.5229e-1 (5.40e-3) - & 3.7495e-1 (6.20e-3) - & \best{3.8297e-1 (6.83e-4)} \\
 &  & 500 & 3.4059e-1 (8.31e-3) - & 2.5026e-1 (3.80e-3) - & 2.7896e-1 (1.85e-3) - & 2.9443e-1 (2.25e-3) - & 3.5913e-1 (3.13e-4) - & 3.6589e-1 (3.04e-3) - & 3.6974e-1 (5.20e-3) - & \best{3.9174e-1 (1.46e-3)} \\
 &  & 800 & 3.4392e-1 (8.80e-3) - & 2.4234e-1 (2.49e-3) - & 2.7689e-1 (2.54e-3) - & 2.9090e-1 (2.25e-3) - & 3.5871e-1 (1.67e-4) - & 3.5966e-1 (1.28e-2) - & 3.5786e-1 (1.50e-2) - & \best{3.9336e-1 (8.58e-4)} \\
 &  & 1000 & 3.5065e-1 (7.58e-3) - & 2.4021e-1 (4.21e-3) - & 2.7526e-1 (2.64e-3) - & 2.9257e-1 (1.57e-3) - & 3.5868e-1 (1.36e-4) - & 3.5993e-1 (8.90e-3) - & 3.5948e-1 (4.74e-3) - & \best{3.9217e-1 (1.05e-3)} \\
\midrule
\multirow{4}{*}{UF5} & \multirow{4}{*}{2} & 300 & 7.1405e-2 (1.33e-2) = & 0.0000e+0 (0.00e+0) - & 0.0000e+0 (0.00e+0) - & 3.7513e-2 (4.02e-2) - & 0.0000e+0 (0.00e+0) - & 7.1357e-2 (5.89e-2) = & 0.0000e+0 (0.00e+0) - & \best{7.9208e-2 (2.28e-2)} \\
 &  & 500 & 8.8698e-2 (3.52e-3) - & 0.0000e+0 (0.00e+0) - & 0.0000e+0 (0.00e+0) - & \best{2.3829e-1 (8.83e-2) +} & 0.0000e+0 (0.00e+0) - & 1.6463e-1 (8.80e-2) = & 9.8271e-2 (1.56e-2) - & 1.8733e-1 (1.30e-2) \\
 &  & 800 & 9.5429e-2 (5.02e-2) - & 0.0000e+0 (0.00e+0) - & 0.0000e+0 (0.00e+0) - & 2.1621e-1 (8.82e-2) - & 0.0000e+0 (0.00e+0) - & 1.9059e-1 (8.07e-2) - & 1.0666e-1 (1.08e-2) - & \best{2.7455e-1 (7.54e-3)} \\
 &  & 1000 & 9.2519e-2 (4.32e-2) - & 0.0000e+0 (0.00e+0) - & 0.0000e+0 (0.00e+0) - & 2.3038e-1 (5.95e-2) - & 0.0000e+0 (0.00e+0) - & 1.3591e-1 (1.10e-1) - & 8.0527e-2 (1.26e-2) - & \best{2.9232e-1 (4.76e-3)} \\
\midrule
\multirow{4}{*}{UF6} & \multirow{4}{*}{2} & 300 & 9.0869e-2 (4.12e-5) - & 0.0000e+0 (0.00e+0) - & 0.0000e+0 (0.00e+0) - & 2.2942e-1 (8.60e-2) - & 0.0000e+0 (0.00e+0) - & 2.6079e-1 (4.17e-2) - & 3.4027e-1 (3.83e-2) - & \best{4.6492e-1 (6.02e-3)} \\
 &  & 500 & 9.0863e-2 (1.51e-4) - & 0.0000e+0 (0.00e+0) - & 0.0000e+0 (0.00e+0) - & 2.8620e-1 (7.23e-2) - & 0.0000e+0 (0.00e+0) - & 2.7395e-1 (6.56e-2) - & 3.3672e-1 (6.08e-2) - & \best{4.9994e-1 (1.36e-2)} \\
 &  & 800 & 9.0904e-2 (5.46e-6) - & 0.0000e+0 (0.00e+0) - & 0.0000e+0 (0.00e+0) - & 2.4141e-1 (7.40e-2) - & 0.0000e+0 (0.00e+0) - & 2.5043e-1 (6.76e-2) - & 3.4596e-1 (4.26e-2) - & \best{5.0978e-1 (2.62e-3)} \\
 &  & 1000 & 9.0912e-2 (1.24e-5) - & 0.0000e+0 (0.00e+0) - & 0.0000e+0 (0.00e+0) - & 2.6901e-1 (8.59e-2) - & 0.0000e+0 (0.00e+0) - & 2.7412e-1 (1.91e-2) - & 3.4676e-1 (3.23e-2) - & \best{5.0883e-1 (1.74e-3)} \\
\midrule
\multirow{4}{*}{UF7} & \multirow{4}{*}{2} & 300 & 4.3207e-1 (2.13e-1) - & 7.2345e-4 (2.49e-3) - & 3.3415e-2 (5.69e-2) - & 4.6187e-1 (1.13e-1) - & 2.4152e-1 (5.75e-2) - & 5.2336e-1 (1.12e-2) - & 5.2793e-1 (2.05e-3) - & \best{5.7686e-1 (2.88e-3)} \\
 &  & 500 & 4.9376e-1 (1.65e-1) - & 1.7222e-4 (3.21e-4) - & 5.3791e-2 (7.03e-2) - & 5.2510e-1 (8.74e-2) - & 2.4646e-1 (4.67e-2) - & 5.3640e-1 (8.00e-3) - & 5.3889e-1 (1.29e-3) - & \best{5.8159e-1 (4.26e-4)} \\
 &  & 800 & 4.0529e-1 (2.30e-1) - & 1.2977e-4 (5.03e-4) - & 4.3283e-2 (5.49e-2) - & 5.4173e-1 (1.81e-2) - & 2.2106e-1 (5.94e-2) - & 5.2644e-1 (9.01e-3) - & 5.3749e-1 (1.17e-3) - & \best{5.8144e-1 (2.78e-4)} \\
 &  & 1000 & 4.0018e-1 (2.27e-1) - & 0.0000e+0 (0.00e+0) - & 6.4905e-2 (5.49e-2) - & 5.1718e-1 (8.91e-2) - & 1.7683e-1 (5.48e-2) - & 5.1210e-1 (1.59e-2) - & 5.3455e-1 (9.10e-4) - & \best{5.8067e-1 (2.85e-4)} \\
\midrule
\multirow{4}{*}{UF8} & \multirow{4}{*}{3} & 300 & 4.2040e-1 (1.34e-1) - & 0.0000e+0 (0.00e+0) - & 6.6463e-2 (1.25e-2) - & 3.3615e-1 (2.59e-3) - & 2.0190e-1 (4.82e-2) - & 3.3508e-1 (3.55e-3) - & 3.2084e-1 (4.05e-3) - & \best{4.8681e-1 (1.35e-2)} \\
 &  & 500 & 4.4862e-1 (9.90e-2) - & 0.0000e+0 (0.00e+0) - & 6.9092e-2 (1.88e-2) - & 3.3188e-1 (1.46e-3) - & 2.5318e-1 (3.46e-2) - & 3.3726e-1 (6.22e-3) - & 3.3289e-1 (2.67e-3) - & \best{5.1341e-1 (1.72e-2)} \\
 &  & 800 & 4.2368e-1 (1.35e-1) = & 0.0000e+0 (0.00e+0) - & 1.1692e-1 (9.78e-2) - & 3.3133e-1 (2.99e-5) - & 2.7829e-1 (3.42e-2) - & 3.3755e-1 (4.83e-3) - & 3.3391e-1 (1.97e-3) - & \best{4.8884e-1 (2.01e-2)} \\
 &  & 1000 & 4.2317e-1 (1.35e-1) = & 0.0000e+0 (0.00e+0) - & 1.0194e-1 (9.63e-2) - & 3.3155e-1 (7.68e-4) - & 2.8890e-1 (2.14e-2) - & 3.3636e-1 (7.00e-3) - & 3.3102e-1 (1.46e-3) - & \best{4.5284e-1 (3.78e-2)} \\
\midrule
\multirow{4}{*}{UF9} & \multirow{4}{*}{3} & 300 & 5.8083e-1 (4.14e-2) - & 0.0000e+0 (0.00e+0) - & 1.2746e-1 (1.67e-2) - & 6.0652e-1 (9.10e-3) - & 2.4282e-1 (1.18e-2) - & 5.1206e-1 (1.89e-2) - & 3.0390e-1 (3.19e-2) - & \best{7.3993e-1 (2.92e-2)} \\
 &  & 500 & 5.7606e-1 (4.25e-2) - & 0.0000e+0 (0.00e+0) - & 1.2834e-1 (1.76e-2) - & 6.4061e-1 (1.75e-2) - & 2.7049e-1 (2.05e-2) - & 5.3229e-1 (4.09e-3) - & 3.6414e-1 (4.23e-2) - & \best{7.7578e-1 (2.24e-2)} \\
 &  & 800 & 6.0747e-1 (7.51e-2) - & 0.0000e+0 (0.00e+0) - & 1.3492e-1 (2.08e-2) - & 6.4506e-1 (1.23e-2) - & 2.6928e-1 (1.32e-2) - & 5.4285e-1 (4.48e-2) - & 3.1441e-1 (3.45e-2) - & \best{7.4760e-1 (1.39e-2)} \\
 &  & 1000 & 5.8562e-1 (5.77e-2) - & 0.0000e+0 (0.00e+0) - & 1.3725e-1 (2.17e-2) - & 6.4552e-1 (1.83e-2) - & 2.7158e-1 (1.67e-2) - & 5.1778e-1 (4.28e-3) - & 2.7372e-1 (1.95e-2) - & \best{7.3614e-1 (1.11e-2)} \\
\midrule
\multirow{4}{*}{UF10} & \multirow{4}{*}{3} & 300 & 4.5741e-2 (1.21e-1) - & 0.0000e+0 (0.00e+0) - & 0.0000e+0 (0.00e+0) - & 2.1406e-2 (1.25e-2) - & 0.0000e+0 (0.00e+0) - & \best{9.3046e-2 (5.09e-2) =} & 5.7075e-2 (3.60e-2) = & 6.9394e-2 (3.45e-2) \\
 &  & 500 & 4.5793e-2 (1.21e-1) - & 0.0000e+0 (0.00e+0) - & 0.0000e+0 (0.00e+0) - & 8.3232e-2 (4.44e-2) - & 0.0000e+0 (0.00e+0) - & 5.5556e-2 (4.47e-2) - & 7.0780e-2 (1.97e-2) - & \best{1.2630e-1 (4.00e-2)} \\
 &  & 800 & \best{1.3730e-1 (1.74e-1) =} & 0.0000e+0 (0.00e+0) - & 0.0000e+0 (0.00e+0) - & 7.8220e-2 (4.45e-2) = & 0.0000e+0 (0.00e+0) - & 2.2268e-2 (2.33e-2) - & 4.0692e-2 (4.91e-2) - & 8.1541e-2 (3.47e-2) \\
 &  & 1000 & 3.0229e-2 (1.74e-2) = & 0.0000e+0 (0.00e+0) - & 0.0000e+0 (0.00e+0) - & 7.4181e-2 (1.76e-2) = & 0.0000e+0 (0.00e+0) - & 1.8305e-2 (4.02e-2) = & 1.0466e-2 (1.81e-2) = & \best{1.3730e-1 (1.74e-1)} \\
\midrule
+/-/= &  &  & 1/31/8 & 0/40/0 & 0/40/0 & 4/34/2 & 0/40/0 & 0/36/4 & 0/38/2 &  \\
\bottomrule
\end{tabular}%
}
\end{table}
\end{landscape}

\clearpage
\twocolumn
\section{Detailed Ablation Results}
\label{supp:ablation}

This section reports the detailed IGD results of the ablation study on the $500$-dimensional test problems. The symbols ``+'', ``-'', and ``='' indicate that the corresponding variant is significantly better than, significantly worse than, and statistically similar to DDFS, respectively. The best mean IGD value for each test problem is highlighted with gray background and bold font.

\clearpage
\onecolumn
\begin{table}[H]
\centering
\caption{Detailed IGD results of the ablation study on 500-dimensional test problems.}
\label{tab:supp_ablation_igd}
\tiny
\setlength{\tabcolsep}{1.0pt}
\renewcommand{\arraystretch}{0.82}
\resizebox{\textwidth}{!}{%
\begin{tabular}{ccc|c|c|c|c|c|c}
\toprule
Problem & M & D & FixedStage & NoSensitivityRefinement & NoStage3Disabling & RandomVariableCategories & UniformFuzzy & DDFS \\
\midrule
LSMOP1 & 2 & 500 & 2.0221e-3 (1.62e-5) - & 1.8658e-3 (1.90e-5) - & 2.5076e-3 (8.72e-5) - & 1.8754e-3 (2.82e-5) - & 1.8477e-3 (4.87e-6) - & \best{1.8251e-3 (6.28e-6)} \\
LSMOP2 & 2 & 500 & 2.3476e-2 (1.78e-4) - & 2.3671e-2 (2.48e-4) - & 2.6817e-2 (2.18e-4) - & 2.3334e-2 (3.43e-4) - & 2.2881e-2 (2.25e-4) - & \best{2.1394e-2 (5.21e-4)} \\
LSMOP3 & 2 & 500 & 7.0713e-1 (9.71e-6) = & 7.0860e-1 (2.74e-6) - & 7.0860e-1 (0.00e+0) - & \best{6.3931e-1 (1.53e-1) =} & 7.0856e-1 (8.21e-5) - & 6.8128e-1 (1.94e-2) \\
LSMOP4 & 2 & 500 & 5.0820e-3 (5.50e-4) - & 3.5335e-3 (3.37e-4) - & 6.1765e-3 (2.55e-4) - & 3.2542e-3 (5.83e-4) - & 3.3743e-3 (2.65e-4) - & \best{2.0529e-3 (4.65e-5)} \\
LSMOP5 & 2 & 500 & 2.6657e-3 (6.40e-5) - & 2.4256e-3 (4.06e-5) - & 3.2561e-3 (1.96e-4) - & 2.4469e-3 (6.75e-5) - & 2.4739e-3 (3.53e-5) - & \best{2.3468e-3 (1.75e-5)} \\
LSMOP6 & 2 & 500 & 6.3362e-1 (1.49e-1) + & 7.4387e-1 (2.92e-4) = & \best{5.6668e-1 (1.60e-1) +} & 7.4405e-1 (3.85e-4) = & 7.4309e-1 (3.41e-4) = & 7.4877e-1 (1.00e-2) \\
LSMOP7 & 2 & 500 & 1.2660e+0 (5.39e-1) = & 1.2108e+0 (6.64e-1) = & \best{6.1243e-1 (4.20e-1) =} & 1.0014e+0 (6.39e-1) = & 1.5874e+0 (9.32e-1) = & 1.0888e+0 (6.10e-1) \\
LSMOP8 & 2 & 500 & 1.2224e-2 (3.97e-3) - & 9.7118e-3 (4.09e-3) - & 1.7508e-2 (1.22e-3) - & 1.2648e-2 (3.06e-3) - & 9.7344e-3 (3.42e-3) - & \best{2.5929e-3 (6.78e-5)} \\
LSMOP9 & 2 & 500 & 7.3735e-1 (1.63e-1) - & 1.4851e-2 (3.58e-3) - & 2.7328e-2 (4.63e-3) - & 1.5505e-2 (5.71e-3) - & 1.7844e-2 (5.59e-3) - & \best{4.8994e-3 (8.10e-5)} \\
UF1 & 2 & 500 & 2.4973e-2 (2.88e-2) - & 1.1319e-2 (1.63e-2) = & 4.6770e-2 (3.51e-2) - & 3.8355e-3 (1.22e-3) = & 4.2727e-3 (1.20e-3) = & \best{3.1997e-3 (5.81e-5)} \\
UF2 & 2 & 500 & 8.3291e-3 (5.98e-4) - & 6.2682e-3 (1.89e-4) = & 1.3818e-2 (9.39e-4) - & 6.6970e-3 (1.27e-3) = & 6.9512e-3 (1.25e-3) = & \best{5.8557e-3 (1.27e-3)} \\
UF3 & 2 & 500 & 4.2158e-3 (4.15e-4) - & \best{2.8993e-3 (4.26e-5) +} & 6.5547e-3 (1.35e-3) - & 2.9074e-3 (1.24e-4) = & 3.1906e-3 (5.07e-4) = & 3.0536e-3 (3.58e-5) \\
UF4 & 2 & 500 & 1.2041e-1 (2.70e-3) - & 1.1967e-1 (1.37e-3) - & 1.2694e-1 (2.35e-3) - & 1.2042e-1 (9.27e-4) - & 1.1925e-1 (1.78e-3) - & \best{3.9378e-2 (1.84e-3)} \\
UF5 & 2 & 500 & 1.0023e+0 (1.06e-1) - & 6.1496e-1 (3.36e-2) - & 9.7162e-1 (1.40e-1) - & 6.0172e-1 (1.06e-1) - & 6.4355e-1 (9.71e-2) - & \best{3.4377e-1 (1.70e-2)} \\
UF6 & 2 & 500 & 3.6333e-1 (8.44e-3) - & 2.8939e-1 (1.49e-1) - & 3.1022e-1 (1.31e-1) - & 2.7544e-1 (1.20e-1) - & 2.5839e-1 (1.79e-1) - & \best{2.7730e-2 (1.42e-2)} \\
UF7 & 2 & 500 & 1.5693e-2 (1.26e-2) = & 3.0439e-2 (1.23e-2) - & 4.5071e-2 (3.07e-3) - & 8.6947e-2 (1.38e-1) - & 3.6743e-2 (9.22e-3) - & \best{3.1840e-3 (2.75e-4)} \\
UF8 & 3 & 500 & 5.0894e-1 (3.15e-2) - & 5.3568e-1 (1.54e-2) - & 5.4189e-1 (1.95e-3) - & 5.0977e-1 (6.69e-2) - & 5.4256e-1 (1.08e-5) - & \best{1.0043e-1 (3.72e-2)} \\
UF9 & 3 & 500 & 3.0485e-1 (7.13e-3) - & 2.9186e-1 (3.35e-2) - & 2.7778e-1 (2.67e-2) - & 2.7997e-1 (3.81e-2) - & 2.9649e-1 (1.76e-3) - & \best{6.1322e-2 (1.56e-2)} \\
UF10 & 3 & 500 & 6.6650e-1 (2.26e-2) - & 6.5839e-1 (1.81e-2) - & 6.8479e-1 (2.16e-2) - & 6.3953e-1 (3.42e-2) - & 6.3661e-1 (5.40e-2) = & \best{5.4304e-1 (6.03e-2)} \\
+/-/= &  &  & 1/15/3 & 1/14/4 & 1/17/1 & 0/13/6 & 0/13/6 &  \\
\bottomrule
\end{tabular}%
}
\end{table}

\clearpage
\onecolumn
\begin{landscape}
\section{Detailed Sensitivity Results}
\label{supp:sensitivity_results}

This section reports the detailed IGD results for different stage-transition and fuzzy-granularity parameter settings on the $500$-dimensional test problems. The symbols ``+'', ``-'', and ``='' have the same meanings as in the main manuscript. The best mean IGD value for each test problem is highlighted with gray background and bold font.

\begin{table}[H]
\vspace{-2mm}
\centering
\caption{Detailed IGD results of the sensitivity analysis on stage-transition parameters.}\vspace{-1mm}
\label{tab:supp_stage_parameter_igd}
\tiny
\setlength{\tabcolsep}{1.0pt}
\renewcommand{\arraystretch}{0.82}
\resizebox{\linewidth}{!}{%
\begin{tabular}{ccc|c|c|c|c|c|c|c}
\toprule
Problem & M & D & $K=2$ & $K=3$ & $\tau_1=0.5$ & $\tau_1=0.7$ & $\tau_2=0.2$ & $\tau_2=0.3$ & DDFS \\
\midrule
LSMOP1 & 2 & 500 & 2.0265e-3 (2.69e-5) - & 1.9781e-3 (3.32e-5) - & 1.9713e-3 (1.84e-5) - & 1.9806e-3 (1.70e-5) - & 2.1873e-3 (5.42e-5) - & 1.9754e-3 (2.10e-5) - & \best{1.9147e-3 (2.37e-5)} \\
LSMOP2 & 2 & 500 & 2.2519e-2 (4.81e-4) - & 2.2303e-2 (2.69e-4) - & 2.2047e-2 (3.67e-4) = & 2.2366e-2 (3.52e-4) - & 2.2980e-2 (3.66e-4) - & 2.2170e-2 (3.53e-4) - & \best{2.1871e-2 (3.60e-4)} \\
LSMOP3 & 2 & 500 & 6.4209e-1 (8.09e-2) + & 6.6324e-1 (7.26e-2) = & 1.4405e+0 (6.29e-1) - & 6.3696e-1 (7.62e-2) + & 6.4449e-1 (9.82e-2) = & 6.5991e-1 (5.14e-2) + & \best{6.3461e-1 (1.09e-1)} \\
LSMOP4 & 2 & 500 & 2.1505e-3 (2.77e-5) - & 2.0801e-3 (3.55e-5) - & 2.0797e-3 (3.03e-5) - & 2.0867e-3 (1.95e-5) - & 2.3247e-3 (1.18e-4) - & 2.0811e-3 (4.34e-5) - & \best{2.0407e-3 (2.46e-5)} \\
LSMOP5 & 2 & 500 & 3.2847e-3 (7.00e-5) - & 3.1713e-3 (6.57e-5) - & 3.1592e-3 (7.67e-5) - & 3.1862e-3 (1.05e-4) - & 3.6271e-3 (1.89e-4) - & 3.1477e-3 (1.12e-4) - & \best{3.0071e-3 (1.15e-4)} \\
LSMOP6 & 2 & 500 & 7.4347e-1 (2.20e-4) + & 7.1876e-1 (7.16e-2) + & 7.6985e-1 (1.99e-2) - & 7.4335e-1 (3.52e-4) + & \best{7.0743e-1 (9.56e-2) +} & 7.0791e-1 (7.74e-2) + & 7.2016e-1 (6.20e-2) \\
LSMOP7 & 2 & 500 & \best{7.9976e-1 (4.75e-1) =} & 1.0431e+0 (5.08e-1) = & 1.0674e+0 (3.79e-1) = & 9.0759e-1 (3.83e-1) = & 9.0933e-1 (5.42e-1) = & 9.3416e-1 (4.62e-1) = & 1.0126e+0 (4.60e-1) \\
LSMOP8 & 2 & 500 & 2.7751e-3 (6.49e-5) - & 2.7105e-3 (5.72e-5) - & 2.7206e-3 (1.11e-4) - & 2.7083e-3 (7.03e-5) - & 3.0785e-3 (1.36e-4) - & 2.7134e-3 (7.93e-5) - & \best{2.5838e-3 (6.17e-5)} \\
LSMOP9 & 2 & 500 & 5.3011e-3 (1.93e-4) - & 5.1016e-3 (1.40e-4) - & 5.0547e-3 (8.33e-5) - & 5.1697e-3 (1.34e-4) - & 5.8941e-3 (2.75e-4) - & 5.1017e-3 (1.21e-4) - & \best{4.7980e-3 (1.32e-4)} \\
UF1 & 2 & 500 & 3.4020e-3 (7.75e-5) - & 3.2904e-3 (5.99e-5) - & 3.2673e-3 (6.31e-5) = & 3.3152e-3 (9.54e-5) - & 3.8931e-3 (9.81e-5) - & 3.3028e-3 (8.75e-5) - & \best{3.2200e-3 (7.55e-5)} \\
UF2 & 2 & 500 & 6.1848e-3 (9.37e-4) - & 5.8657e-3 (8.13e-4) = & 5.9366e-3 (8.56e-4) = & 6.1566e-3 (8.15e-4) - & 6.5125e-3 (9.37e-4) - & 5.5286e-3 (6.78e-4) = & \best{5.4037e-3 (5.52e-4)} \\
UF3 & 2 & 500 & 3.2769e-3 (8.08e-5) - & 3.1517e-3 (7.67e-5) = & 3.2145e-3 (3.82e-4) = & 3.1485e-3 (8.31e-5) = & 3.6622e-3 (1.73e-4) - & 3.1882e-3 (1.31e-4) = & \best{3.0948e-3 (1.02e-4)} \\
UF4 & 2 & 500 & 4.1311e-2 (1.62e-3) - & 4.0371e-2 (1.14e-3) - & 4.0151e-2 (1.17e-3) - & 4.0736e-2 (1.07e-3) - & 4.2959e-2 (1.29e-3) - & 3.9954e-2 (1.03e-3) - & \best{3.8748e-2 (8.43e-4)} \\
UF5 & 2 & 500 & 3.3997e-1 (1.51e-2) - & 3.2954e-1 (1.83e-2) = & 3.3366e-1 (1.52e-2) = & 3.2972e-1 (1.60e-2) = & 3.3942e-1 (1.42e-2) - & 3.3638e-1 (1.51e-2) - & \best{3.2335e-1 (1.61e-2)} \\
UF6 & 2 & 500 & 1.9698e-2 (9.58e-3) - & 1.6717e-2 (7.44e-3) = & 1.8714e-2 (1.03e-2) = & 1.7221e-2 (9.02e-3) = & 2.1194e-2 (6.20e-3) - & 1.7234e-2 (6.85e-3) = & \best{1.4867e-2 (2.37e-3)} \\
UF7 & 2 & 500 & 3.2323e-3 (1.49e-4) = & 3.1914e-3 (1.68e-4) = & 3.1641e-3 (1.60e-4) = & 3.2681e-3 (1.64e-4) - & 4.0069e-3 (1.87e-4) - & 3.2432e-3 (1.80e-4) = & \best{3.1395e-3 (1.58e-4)} \\
UF8 & 3 & 500 & 1.4411e-1 (1.15e-1) = & 1.8062e-1 (1.51e-1) = & 1.3694e-1 (1.25e-1) = & 1.2326e-1 (1.22e-1) = & 1.1390e-1 (4.48e-2) = & 1.3475e-1 (5.14e-2) = & \best{9.4975e-2 (4.19e-2)} \\
UF9 & 3 & 500 & 5.7171e-2 (1.75e-2) = & 6.0621e-2 (1.29e-2) = & 5.6800e-2 (1.16e-2) = & 4.9528e-2 (1.26e-2) = & 6.0750e-2 (1.39e-2) = & 5.4569e-2 (9.97e-3) = & \best{4.9026e-2 (1.22e-2)} \\
UF10 & 3 & 500 & 4.9340e-1 (5.50e-2) = & 5.2433e-1 (7.53e-2) = & 8.9792e-1 (3.70e-1) - & 4.9420e-1 (7.57e-2) = & \best{4.7350e-1 (5.90e-2) =} & 5.0035e-1 (4.64e-2) = & 4.8182e-1 (5.18e-2) \\
+/-/= &  &  & 2/12/5 & 1/8/10 & 0/9/10 & 2/10/7 & 1/13/5 & 2/9/8 &  \\
\bottomrule
\end{tabular}%
}
\end{table}

\vspace{-3mm}
\begin{table}[H]
\vspace{-2mm}
\centering
\caption{Detailed IGD results of the sensitivity analysis on fuzzy-granularity parameters.}\vspace{-1mm}
\label{tab:supp_fuzzy_parameter_igd}
\tiny
\setlength{\tabcolsep}{1.0pt}
\renewcommand{\arraystretch}{0.82}
\resizebox{\linewidth}{!}{%
\begin{tabular}{ccc|c|c|c|c|c|c|c|c|c}
\toprule
Problem & M & D & $I_{\mathrm{div}}=5/50$ & $I_{\mathrm{div}}=20/200$ & $I_{\mathrm{div}}=50/500$ & $I_{\mathrm{ch}}=500/5000$ & $I_{\mathrm{ch}}=2000/20000$ & $I_{\mathrm{cl}}=50/500$ & $I_{\mathrm{cl}}=200/2000$ & $I_{\mathrm{cl}}=500/5000$ & DDFS \\
\midrule
LSMOP1 & 2 & 500 & 1.9184e-3 (1.70e-5) = & 1.9182e-3 (1.54e-5) = & \best{1.9129e-3 (1.40e-5) =} & 1.9185e-3 (1.35e-5) = & 1.9302e-3 (2.32e-5) = & 1.9259e-3 (1.85e-5) = & 1.9258e-3 (1.89e-5) = & 1.9352e-3 (1.61e-5) = & 1.9199e-3 (1.85e-5) \\
LSMOP2 & 2 & 500 & \best{2.1368e-2 (5.33e-4) +} & 2.2196e-2 (3.10e-4) = & 2.2221e-2 (5.39e-4) = & 2.1789e-2 (2.87e-4) = & 2.1995e-2 (5.03e-4) = & 2.2176e-2 (6.01e-4) = & 2.2236e-2 (3.35e-4) - & 2.1804e-2 (1.49e-4) = & 2.1883e-2 (3.28e-4) \\
LSMOP3 & 2 & 500 & 6.9624e-1 (1.95e-2) = & \best{6.5370e-1 (9.34e-2) =} & 6.5913e-1 (4.84e-2) = & 6.8822e-1 (1.84e-2) = & 6.7140e-1 (6.19e-2) = & 6.8411e-1 (1.76e-2) - & 6.8672e-1 (2.16e-2) = & 6.7732e-1 (1.46e-2) = & 6.7871e-1 (1.78e-2) \\
LSMOP4 & 2 & 500 & 2.0344e-3 (2.50e-5) = & \best{2.0202e-3 (2.10e-5) =} & 2.0413e-3 (3.82e-5) = & 2.0211e-3 (2.65e-5) = & 2.0273e-3 (3.01e-5) = & 2.0224e-3 (3.54e-5) = & 2.0533e-3 (6.08e-5) = & 2.0400e-3 (4.28e-5) = & 2.0361e-3 (2.96e-5) \\
LSMOP5 & 2 & 500 & 3.1190e-3 (1.17e-4) = & 3.0211e-3 (1.11e-4) = & \best{2.9916e-3 (8.16e-5) =} & 3.0931e-3 (6.34e-5) = & 3.0462e-3 (8.67e-5) = & 3.0130e-3 (6.67e-5) = & 3.0568e-3 (1.30e-4) = & 3.0668e-3 (6.27e-5) = & 3.0737e-3 (8.02e-5) \\
LSMOP6 & 2 & 500 & 7.6656e-1 (4.47e-3) - & 7.3242e-1 (5.92e-2) + & \best{7.2823e-1 (5.20e-2) +} & 7.3394e-1 (8.59e-2) = & 7.6320e-1 (4.28e-3) = & 7.5932e-1 (1.09e-2) = & 7.6442e-1 (4.46e-3) - & 7.6308e-1 (7.12e-3) = & 7.3700e-1 (6.97e-2) \\
LSMOP7 & 2 & 500 & 1.2272e+0 (2.77e-1) - & 9.6139e-1 (5.75e-1) = & 9.2660e-1 (5.97e-1) = & 1.0129e+0 (2.97e-1) = & 1.0650e+0 (3.86e-1) = & 1.0257e+0 (2.76e-1) = & 1.0042e+0 (5.01e-1) = & 1.2209e+0 (2.05e-1) = & \best{8.7566e-1 (2.91e-1)} \\
LSMOP8 & 2 & 500 & 2.7048e-3 (6.27e-5) = & 2.6408e-3 (8.10e-5) = & 2.6368e-3 (5.64e-5) = & 2.6382e-3 (4.55e-5) = & 2.6383e-3 (8.12e-5) = & 2.6481e-3 (6.03e-5) = & \best{2.6158e-3 (5.61e-5) =} & 2.6579e-3 (6.79e-5) = & 2.6581e-3 (7.45e-5) \\
LSMOP9 & 2 & 500 & 4.9718e-3 (5.02e-5) = & 4.8716e-3 (9.81e-5) = & 4.8569e-3 (9.67e-5) = & 4.8775e-3 (1.32e-4) = & 4.8561e-3 (4.31e-5) = & 4.9032e-3 (1.17e-4) = & 4.8494e-3 (1.46e-4) = & \best{4.8369e-3 (7.78e-5) +} & 4.9385e-3 (8.85e-5) \\
UF1 & 2 & 500 & 3.2211e-3 (8.78e-5) = & 3.1923e-3 (6.10e-5) = & 3.2118e-3 (5.86e-5) = & 3.2086e-3 (9.68e-5) = & 3.2647e-3 (7.49e-5) = & 3.2007e-3 (4.40e-5) = & \best{3.1857e-3 (7.34e-5) =} & 3.2215e-3 (5.02e-5) = & 3.1965e-3 (7.40e-5) \\
UF2 & 2 & 500 & 5.6664e-3 (7.42e-4) = & 5.7649e-3 (7.85e-4) = & 6.3867e-3 (9.29e-4) = & \best{5.5482e-3 (6.56e-4) =} & 5.7548e-3 (7.52e-4) = & 5.7044e-3 (6.44e-4) = & 5.8825e-3 (9.89e-4) = & 5.5789e-3 (6.19e-4) = & 5.9123e-3 (8.75e-4) \\
UF3 & 2 & 500 & \best{3.0673e-3 (7.15e-5) =} & 3.1286e-3 (8.60e-5) = & 3.1201e-3 (7.83e-5) = & 3.0980e-3 (1.59e-4) = & 3.0742e-3 (3.49e-5) = & 3.0919e-3 (6.97e-5) = & 3.0883e-3 (7.72e-5) = & 3.1054e-3 (1.09e-4) = & 3.1230e-3 (9.18e-5) \\
UF4 & 2 & 500 & 3.8888e-2 (6.74e-4) + & 3.9269e-2 (1.29e-3) = & 4.0502e-2 (1.83e-3) = & 3.9544e-2 (1.64e-3) = & 3.9704e-2 (1.53e-3) = & 3.9827e-2 (1.26e-3) = & 3.9536e-2 (1.65e-3) = & \best{3.8777e-2 (8.95e-4) +} & 3.9854e-2 (1.04e-3) \\
UF5 & 2 & 500 & 3.2814e-1 (1.16e-2) = & 3.3382e-1 (1.79e-2) = & 3.3221e-1 (1.35e-2) = & 3.3532e-1 (9.96e-3) = & 3.2901e-1 (2.22e-2) = & 3.3307e-1 (1.94e-2) = & 3.3884e-1 (1.69e-2) = & \best{3.1974e-1 (2.16e-2) =} & 3.3378e-1 (1.99e-2) \\
UF6 & 2 & 500 & 1.6063e-2 (1.82e-3) = & 2.4144e-2 (1.52e-2) = & 1.6568e-2 (4.34e-3) = & \best{1.5930e-2 (2.07e-3) =} & 1.8476e-2 (5.36e-3) = & 1.8864e-2 (9.61e-3) = & 2.1583e-2 (1.09e-2) = & 2.6375e-2 (1.39e-2) = & 1.9691e-2 (9.57e-3) \\
UF7 & 2 & 500 & 3.2119e-3 (1.54e-4) = & 3.1370e-3 (2.93e-4) = & \best{3.1224e-3 (1.46e-4) =} & 3.2257e-3 (2.57e-4) = & 3.2553e-3 (3.08e-4) = & 3.1350e-3 (1.99e-4) = & 3.2247e-3 (1.94e-4) = & 3.2620e-3 (2.16e-4) = & 3.2146e-3 (2.40e-4) \\
UF8 & 3 & 500 & 2.3332e-1 (2.17e-1) = & 1.2968e-1 (6.70e-2) = & \best{1.1398e-1 (5.91e-2) =} & 2.0065e-1 (1.93e-1) = & 1.2304e-1 (6.64e-2) = & 1.8809e-1 (1.94e-1) = & 1.9735e-1 (1.84e-1) = & 1.3452e-1 (6.49e-2) = & 1.1633e-1 (5.10e-2) \\
UF9 & 3 & 500 & 5.8585e-2 (1.66e-2) = & 6.1805e-2 (1.66e-2) = & 5.0199e-2 (1.21e-2) = & 5.4615e-2 (1.52e-2) = & 5.2405e-2 (1.78e-2) = & 5.2688e-2 (1.22e-2) = & 5.4143e-2 (1.11e-2) = & \best{4.9816e-2 (1.40e-2) =} & 5.7797e-2 (6.07e-3) \\
UF10 & 3 & 500 & 4.9313e-1 (8.06e-2) = & 5.0238e-1 (4.82e-2) = & 4.9849e-1 (4.49e-2) = & \best{4.7382e-1 (7.17e-2) =} & 4.9023e-1 (6.07e-2) = & 4.9536e-1 (8.56e-2) = & 5.2450e-1 (4.28e-2) = & 5.1519e-1 (3.78e-2) = & 5.0661e-1 (7.10e-2) \\
+/-/= &  &  & 2/2/15 & 1/0/18 & 1/0/18 & 0/0/19 & 0/0/19 & 0/1/18 & 0/2/17 & 2/0/17 &  \\
\bottomrule
\end{tabular}%
}
\end{table}

\end{landscape}

\end{document}